\documentclass[lettersize,journal]{IEEEtran}

\usepackage{amsmath,amsfonts}
\usepackage{algorithmic}
\usepackage{algorithm}
\usepackage{array}
\usepackage[caption=false,font=normalsize,labelfont=sf,textfont=sf]{subfig}
\usepackage{textcomp}
\usepackage{stfloats}
\usepackage{url}
\usepackage{verbatim}
\usepackage{graphicx}
\usepackage{cite}

\usepackage{xcolor}
\usepackage{orcidlink}
\usepackage[misc]{ifsym} %

\usepackage{booktabs}
\usepackage{multirow}
\usepackage{hyperref}
\usepackage{fontawesome5}
\usepackage{makecell}

\usepackage{amsmath,amsfonts,bm}

\def\eqref#1{equation~\ref{#1}}

\def\1{\bm{1}}

\DeclareMathAlphabet{\mathsfit}{\encodingdefault}{\sfdefault}{m}{sl}
\SetMathAlphabet{\mathsfit}{bold}{\encodingdefault}{\sfdefault}{bx}{n}

\newcommand{\Imat}{{\bf I}}

\newcommand{\hflogo}{\includegraphics[height=0.8em]{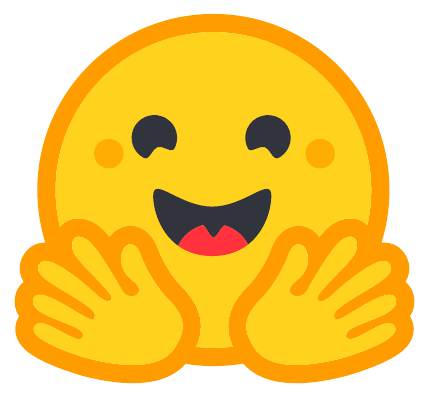}}
\def\githuburl{https://github.com/VisionXLab/RSCoVLM}
\def\huggingfaceurl{https://huggingface.co/datasets/Qingyun/remote-sensing-sft-data}

\begin{document}

\title{
    Co-Training Vision Language Models \\for Remote Sensing Multi-task Learning
}

\author{
    \IEEEauthorblockN{
        Qingyun Li$^{1*}$\orcidlink{0000-0001-5101-4937}, 
        Shuran Ma$^{3*}$, 
        Junwei Luo$^{5*}$, 
        Yi Yu$^{6*}$,
        Yue Zhou$^{4\ddagger}$\orcidlink{0000-0002-3080-6721},
        Fengxiang Wang$^{7}$,
        Xudong Lu$^{8}$,\\
        Xiaoxing Wang$^{2}$,
        Xin He$^{1}$,
        Yushi Chen$^{1}$\textsuperscript{\Letter}\orcidlink{0000-0003-2421-0996}, \textit{Member, IEEE},
        Xue Yang$^{2\ddagger}$\orcidlink{0000-0002-7084-9101}, \textit{Member, IEEE}
    }\\

    \vspace{4px}

    \IEEEauthorblockA{
        \textit{$^1$Harbin Institute of Technology}~
        \textit{$^2$Shanghai Jiao Tong University}~
        \textit{$^3$Xidian University}\\
        \textit{$^4$East China Normal University}~
        \textit{$^5$Wuhan University}~
        \textit{$^6$Southeast University}\\
        \textit{$^7$National University of Defense Technology}~
        \textit{$^8$Chinese University of Hong Kong}
    }\\

    \vspace{4px}

    \IEEEauthorblockA{* Equal Contribution \quad \textsuperscript{\Letter} Corresponding Author \quad $^{\ddagger}$ Project Leader}\\

    \vspace{4px}
    
    \IEEEauthorblockA{\href{\githuburl}{\faGithub}~Code: \url{\githuburl}}\\
    \IEEEauthorblockA{\href{\huggingfaceurl}{\hflogo}~Data: \url{\huggingfaceurl}}

    \thanks{This work was supported by National Natural Science Foundation of China under the Grant 62371169 and 62506229, and Natural Science Foundation of Shanghai under 25ZR1402268.}%
    \thanks{Q. Li, X. He, and Y. Chen are with the School of Electronics and Information Engineering, Harbin Institute of Technology, Harbin 150001, China (e-mails: \url{21B905003@stu.hit.edu.cn}; \url{hexin1@hit.edu.cn}; \url{chenyushi@hit.edu.cn}). \textit{(Corresponding author: Yushi Chen.)}}
    \thanks{S. Ma is with the School of Telecommunications Engineering, Xidian University, Xi'an 710071, China (e-mails: \url{shrma@stu.xidian.edu.cn}).}
    \thanks{J. Luo is with the School of Remote Sensing and Information Engineering, Wuhan University, Wuhan 430072, China (e-mail: \url{luojunwei@whu.edu.cn}).}
    \thanks{Y. Yu is with the School of Automation, Southeast University, Nanjing 210096, China (e-mail: \url{yuyi@seu.edu.cn}).}
    \thanks{F. Wang is with the College of Computer Science and Technology, National University of Defense Technology, Changsha 410005, China (e-mail: \url{wfx23@nudt.edu.cn}).}
    \thanks{X. Lu is with the Department of Electronic Engineering, The Chinese University of Hong Kong, Hong Kong SAR 999077, China (e-mail: \url{luxudong@link.cuhk.edu.hk}.}
    \thanks{X. Wang is with the School of Computer Science, Shanghai Jiao Tong University, Shanghai 200240, China (e-mail: \url{figure1_wxx@sjtu.edu.cn}).}
    \thanks{Y. Zhou is with the School of Geospatial Artificial Intelligence, East China Normal University, Shanghai 200241, China (e-mail: \url{yzhou@geoai.ecnu.edu.cn}).}
    \thanks{X. Yang is with the School of Automation and Intelligent Sensing, Shanghai Jiao Tong University, Shanghai 200240, China (e-mail: \url{yangxue-2019-sjtu@sjtu.edu.cn}).}
\vspace{-25pt}
}

\markboth{\href{https://www.mdpi.com/journal/remotesensing}{Remote Sensing} 2026, 18(2), 222; \href{https://doi.org/10.3390/rs18020222}{DOI: 10.3390/rs18020222}}%
{Shell \MakeLowercase{\textit{et al.}}: A Sample Article Using IEEEtran.cls}

\maketitle

\begin{abstract}
With Transformers achieving outstanding performance on individual remote sensing (RS) tasks, we are now approaching the realization of a unified model that excels across multiple tasks through multi-task learning (MTL).
Compared to single-task approaches, MTL methods offer improved generalization, enhanced scalability, and greater practical applicability.
Recently, vision language models (VLMs) have achieved promising results in RS image understanding, grounding, and ultra-high-resolution (UHR) image reasoning, respectively. Moreover, the unified text-based interface demonstrates significant potential for MTL.
Hence, in this work, we present RSCoVLM, a simple yet flexible VLM baseline for RS MTL.
Firstly, we create the data curation engine, including data acquisition, offline processing and integrating, as well as online loading and weighting. This data engine effectively addresses complex RS data enviroment and generates flexible vision-language conversations.
Furthermore, we propose a unified dynamic-resolution strategy to address the diverse image scales inherent in RS imagery. For UHR images, we introduce the Zoom-in Chain mechanism together with its corresponding dataset, LRS-VQA-Zoom. The strategies are flexible and effectively mitigate the computational burdens.
Additionally, we significantly enhance the model’s object detection capability and propose a novel evaluation protocol that ensures fair comparison between VLMs and conventional detection models. 
Extensive experiments demonstrate that RSCoVLM achieves state-of-the-art performance across diverse tasks, outperforming existing RS VLMs and even rivaling specialized expert models. 
All the training and evaluating tools, model weights, and datasets have been fully open-sourced to support reproducibility. 
We expect that this baseline will promote further progress toward general-purpose RS models.
\end{abstract}


\section{Introduction}

\IEEEPARstart{E}{ARTH} observation systems have acquired extensive remote sensing (RS) data, necessitating the development of automated RS image interpretation techniques~\cite{GRSM22review}. The emergence of artificial general intelligence has inspired researchers in the RS community to develop versatile agents capable of performing multiple tasks, such as scene classification, visual question answering, and object detection~\cite{zhou2024vlgfm}.

Most RS image processing methods typically train a specifically-designed model on isolated datasets to achieve optimal performance on individual tasks. Due to the heterogeneity of data and model architecture, developing a unified model capable of handling multiple RS tasks, i.e., multi-task learning (MTL), remains challenging~\cite{mtl_survey_2022}. 

MTL provides several advantages for RS applications. First, a single MTL model with shared parameters can handle multiple tasks at once, unlike traditional task-specific models, which is closer to human perception. Second, by sharing knowledge across tasks, MTL mitigates the shortage of annotated data and reduces overfitting on individual tasks. Third, MTL learns joint representations that capture correlations among tasks, improving generalization. RS foundation models also benefit by obtaining consistent representations through pre-training on upstream tasks and fine-tuning on various downstream tasks. Overall, MTL helps advance RS foundation models by expanding pre-training tasks and enhancing cross-task learning~\cite{li2024co}.

Transformer~\cite{Transformer} has demonstrated remarkable flexibility and generalization capabilities across various domains, including computer vision~\cite{ViT}, natural language processing, speech processing, and remote sensing data analysis~\cite{TRD}. This progress has brought the goal of a unified multimodal and multi-task architecture increasingly within reach~\cite{han2023onellm}. 
Consequently, vision language models (VLMs), that bridge the gap between the two modalities by learning from vast amounts of paired~\cite{clip} and interleaved image-text data~\cite{li2024omnicorpus}, have been proposed and become the most commonly adopted foundation model paradigm in the multimodal domain~\cite{flamingo}.

In this study, we focus on generative VLMs, also named multimodal language models. These models are typically constructed upon vision and language foundation models, enabling them to process visual inputs and effectively interpret textual instructions. 
By harnessing the capabilities of powerful pre-trained foundation models and leveraging a versatile text interface, VLMs are positioned as a crucial element in the progression toward the unified MTL~\cite{internvl}.

We consider that VLMs represent an ideal paradigm for RS MTL. 
Firstly, the textual interface of VLMs provides a unified representation for diverse task objectives, because the outputs of different RS tasks, such as classification, grounding, captioning, or question answering, can all be expressed in text form. 
Secondly, instruction tuning has demonstrated that VLMs can generalize beyond the tasks seen during training~\cite{llava}, enabling them to handle novel or composite tasks through in-context learning~\cite{flamingo}. 
Finally, with sufficiently strong foundational capabilities, VLMs offer the potential to evolve toward more autonomous RS agents, where task reasoning and workflow design can be accomplished within a single, coherent framework.

\begin{figure}[tb]
  \centering
  \includegraphics[width=\linewidth]{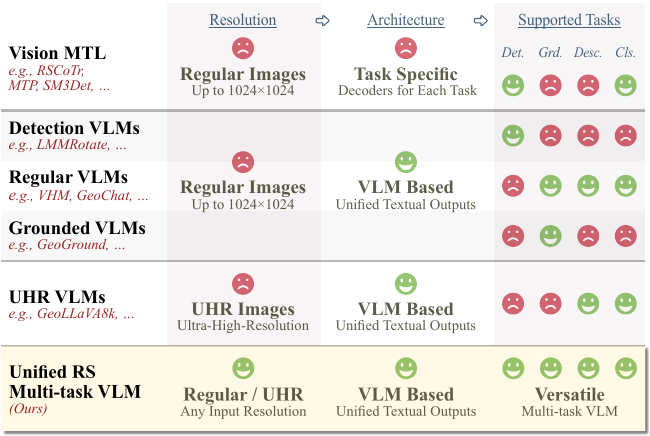}
  \vspace{-5mm}
  \caption{Comparisons with existing MTL methods across the resolutions of input images, network architectures, and supported tasks (i.e., detection, grounding, description, and classification).}
  \label{fig:intro}
  \vspace{-5mm}
\end{figure}

In the RS community, MTL has been preliminarily explored, including several attempts leveraging VLMs. Nevertheless, existing approaches still exhibit notable limitations. Fig.~\ref{fig:intro} summarizes the key differences among representative paradigms.

Early RS MTL approaches were typically designed for multiple pure-vision tasks~\cite{li2024co,MTP,li2024sm3det}, such as classification, segmentation, and detection. These methods generally adopt a shared feature extraction backbone with task-specific output heads. With carefully crafted training strategies, their performance on individual benchmarks is comparable to that of expert models trained on the specific dataset. However, they suffer from limited scalability and architectural rigidity. As the number of tasks increases, the heterogeneous design of multiple heads makes optimization increasingly difficult and less robust. Consequently, this paradigm struggles to scale up, resulting in insufficient model generalization. Nevertheless, when deployed on resource-constrained platforms such as satellites, this kind of MTL model remains highly valuable for its computational and storage efficiency.

As general-purpose VLMs increasingly exhibit early signs of the universal model, they have emerged as a scalable paradigm for MTL. In the RS domain, several studies have explored VLM-based MTL. However, their investigations into unified and generalizable paradigms remain limited: The regular VLMs focus primarily on language-centric description tasks such as image captioning and visual question answering, where text descriptions are synthesized for RS images to enable semantic understanding~\cite{vhm,geochat,lhrsbot}. Others extend VLMs to purely visual tasks such as visual grounding and object detection, leveraging the flexible language interface of VLMs to learn from abundant localization annotations and achieve precise detection capabilities~\cite{geoground,lmmrotate}. In addition, several approaches target ultra-high-resolution (UHR) RS image reasoning, often employing token pruning to alleviate the computational burden caused by extremely large inputs~\cite{lrsvqa,geollava8k}.

These studies collectively highlight the great potential of VLMs for RS MTL, yet each remains constrained within a limited scope. 
As shown in Fig.~\ref{fig:intro}, the first four types of works focus mainly on tasks involving regular images (images with regular resolutions)~\cite{vhm,geochat,lhrsbot,geoground,lmmrotate}, whereas the fifth is specialized for UHR scenarios~\cite{lrsvqa,geollava8k}. The detection VLMs~\cite{lmmrotate} and grounded VLMs~\cite{geoground} excel at spatial grounding but pays little attention to semantic understanding, while the regular VLMs~\cite{vhm,geochat,lhrsbot} rarely explore the crucial object detection capabilities which is essential for RS image analysis. 
Hence, a unified framework that addresses these limitations in an integrated MTL setting is still lacking.

In this paper, we present a novel foundation model named RSCoVLM (\textbf{R}emote \textbf{S}ensing \textbf{Co}operatively-trained \textbf{V}ision \textbf{L}anguage \textbf{M}odel). 
We cooperatively train (co-train) it for multiple tasks in a unified framework, that handles the following problems. 

Firstly, the large-scale multi-task data must be curated to enable effective MTL. However, RS data are inherently complex, often exhibiting inconsistencies in format, noisy annotations, and heterogeneous bounding box definitions. Therefore, careful data curation is required to construct a well-organized and sustainable data enviroment for model training.

Secondly, we need to address the challenge of diverse input sizes of RS images. The classification task often uses size with a few megapixels, such as 256$\times$256. Common object detection models typically use input sizes such as 512$\times$512, 800$\times$800, or 1024$\times$1024. However, UHR images can have widths and heights exceeding 4,000 pixels. Therefore, a dynamic resolution strategy is required, along with efficient and highly compatible solutions for UHR scenarios.

Finally, previous VLMs have shown limited capability in object detection tasks. They either perform only sparse visual grounding~\cite{geochat,geoground}, provide detection results for a single category~\cite{earthgpt}, or are evaluated leniently under low IoU thresholds~\cite{luo2024skysensegpt}. However, aerial detection is particularly challenging due to issues such as dense object distribution, which places high demands on the visual input resolution and output sequence length of MLLMs. Moreover, VLMs cannot directly output the confidence scores of predicted objects, making it difficult to fairly compare them with traditional models using commonly-used evaluation metrics.

To make RSCoVLM a competitive MTL baseline, we address the aforementioned challenges, respectively.
We firstly create a data curation engine, compromising the acquisition of raw data, the offline processing and integration, as well as online loading and weighting. 
Moreover, we propose a dynamic resolution strategy that enables the model to simultaneously learn from images of various sizes. To further enhance the reasoning performance on UHR images, we propose the Zoom-in Chain strategy, which mimics how humans reason over UHR images. We also construct a corresponding dataset, LRS-VQA-Zoom, to specifically strengthen this capability.
Additionally, we apply VLMs to object detection and propose a fair evaluation method that does not rely on confidence thresholds. Based on this, our RSCoVLM is validated as the first VLM that achieves performance comparable to traditional models on dense aerial detection task.

We evaluate RSCoVLM on multiple tasks across various benchmarks, achieving state-of-the-art performance in all of them. Our unified MTL framework greatly improves the model’s generalization ability, scalability, and usability.

To ensure transparency and reproducibility, we fully open-source all details of this work, including the codes, model weights, data folder. We will continuously maintain the open-source resources and update them with our latest research progress, aiming to build a user-friendly platform for the community.

The main contributions are summarized as follows:

\begin{enumerate}
	\item{We present RSCoVLM, a fully open-sourced VLM baseline for RS MTL. The experiment show that our model achieves leading performance across benchmarks of various datasets and tasks. }
	\item{We develop the universal framework for RS MTL based on VLM and create the data curation engine to facilitate unified training across multi datasets of various RS tasks.}
	\item{We proposed a dynamic resolution strategy for RS, along with the Zoom-in Chain strategy and the LRS-VQA-Zoom dataset to further enhance the model’s reasoning ability on UHR images.}
	\item{We develop the auto-regressive aerial detection method for RS VLMs and propose an evaluation metric that enables a fair comparison between RS VLM and conventional methods.}
\end{enumerate}

This manuscript is an extended and improved version of our conference paper~\cite{lmmrotate} published in IGARSS 2025, which only investigate VLMs for detection tasks. 
The autoregressive object detection scheme in Section~\ref{sec:detection} is primarily derived from the conference version. Building upon it, we not only refine detection details in Section~\ref{subsec:detection-method} but also further upgrade the VLMs with unified multi-task learning, accompanied by additional methods, models, and experimental results.

\section{Related Works}

\subsection{General-purpose Vision Language Models}
Early works such as VisualGPT~\cite{visualgpt}, BLIP-2~\cite{blip2}, and Flamingo~\cite{flamingo} explored different ways of integrating visual features with large language models or training joint image-text encoders, showing improved multimodal reasoning and understanding.

Later instruction-tuned frameworks, including LLaVA~\cite{llava}, MiniGPT-4~\cite{zhu2024minigpt}, and InstructBLIP~\cite{InstructBlip}, further enhanced interactive comprehension by fine-tuning LLMs with visual–text instructions. Lightweight adaptation methods such as LLaMA-Adapter~V2~\cite{lamaadapterv2} and SPHINX~\cite{sphinx} improved efficiency through visual adapters and zero-shot attention fusion, reducing the cost of multimodal alignment.

In parallel, the VisionLLM series~\cite{VisionLLM, visionllm-v2} unified vision-centric tasks under the LLM paradigm, enabling open-ended reasoning over diverse visual inputs. More recent large-scale MLLMs, including PaLI-X~\cite{palix}, MiMo-VL~\cite{mimovl}, InternVL~\cite{internvl}, CogVLM~\cite{cogvlm}, and the Qwen-VL series~\cite{qwen2vl, qwen25vl}, further integrate vision and language through end-to-end pretraining on massive multimodal data and scalable architectures. These models show improved visual grounding, OCR, and cross-domain reasoning, representing a shift from adapter-based fusion toward deeply coupled vision-language modeling. Collectively, these advances lay the foundation for adapting MLLMs to specialized domains such as remote sensing, where complex visual semantics and open-ended reasoning are required.

\subsection{Remote Sensing Vision Language Models}
Recently, integrating vision–language models into remote sensing (RS) has attracted growing attention, giving rise to several domain-specific VLMs. GeoChat~\cite{geochat} pioneered this direction as the first RS-oriented VLM, addressing multiple optical imagery tasks via conversational interaction. EarthGPT~\cite{earthgpt} introduced a unified multimodal framework for multi-source RS data and diverse vision-language tasks. LHRS-Bot~\cite{lhrsbot} leverages multi-level vision-language alignment and curriculum learning for RS image understanding.

Beyond static image interpretation, recent works explore temporal and fine-grained understanding. TEOChat~\cite{irvin2024teochat} supports temporal Earth observation imagery and instruction following over sequential frames. SkySenseGPT~\cite{luo2024skysensegpt} extends instruction tuning to fine-grained RS comprehension, achieving strong performance on public datasets and complex comprehension tasks. VHM~\cite{vhm} demonstrates capabilities on tasks such as building vectorization, multi-label classification, and honest question answering.

\subsection{Remote Sensing Multi-task Learning}
While recent advancements in remote sensing VLMs have enabled versatile multi-task capabilities, their general-purpose architectures often implicitly handle task interdependencies, potentially overlooking the intrinsic challenges of multi-task optimization, such as task interference and imbalance. 

One line of research focuses on leveraging shared representations for synergistic task pairing. For example, several studies~\cite{MTSCD-Net, stgnet, SMNet, mcenet} combine semantic segmentation with change detection from bi-temporal imagery, showing that joint learning enhances feature sharing and reduces redundancy. Another direction jointly models geometric and semantic information, such as height estimation with semantic segmentation~\cite{lu2022multi, carvalho2019multitask}, demonstrating gains over single-task baselines.

Beyond specific task pairs, generalized MTL frameworks have been proposed. RSCoTr~\cite{li2024co} performs classification, segmentation, and detection simultaneously, illustrating the potential of unified RS analysis. Large-scale datasets like SatlasPretrain~\cite{Bastani_2023_ICCV} with multiple annotation modalities facilitate advanced MTL model development. SM3Det~\cite{li2024sm3det} uses a mixture-of-experts structure for multi-modal detection of horizontal and rotated bounding boxes. EarthDial~\cite{soni2025earthdial} leverages multiple multi-task decoders to transfer knowledge across diverse tasks, enriching shared feature learning.

\section{Method}

\subsection{The Universal RS Multi-task Framework}

\label{sec:framework}

\begin{figure*}[ht]
    \centering
    \includegraphics[width=\linewidth]{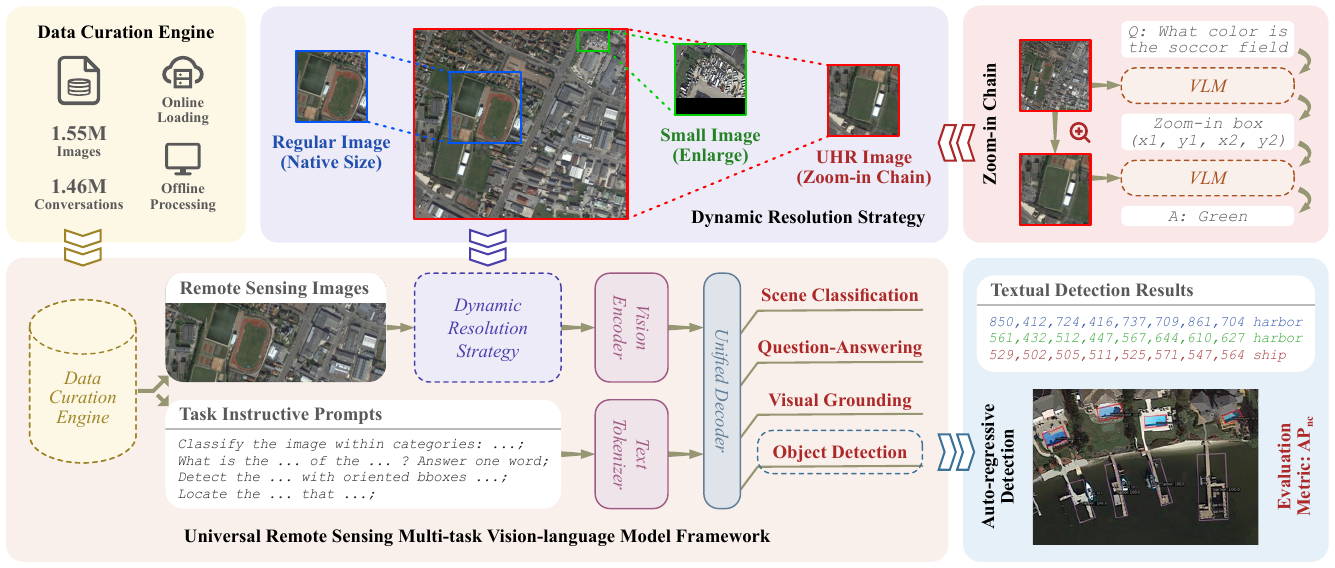}
   \vspace{-5mm}
    \caption{Overall schematic diagram of the proposed method. The overall RS MTL framework based on VLM is presented in Section~\ref{sec:framework}. The data curation engine is introduced in Section~\ref{sec:data}. The dynamic resolution strategy is proposed in Section~\ref{sec:resolution}. We introduce the proposed Zoom-in Chain strategy and the corresponding LRS-VQA-Zoom dataset in Section~\ref{zoomin}. Finally, we describe the aerial detection scheme and propose the fair metric $\text{AP}_{\text{nc}}$ in Section~\ref{sec:detection}.}
    \vspace{-5mm}
    \label{fig:framework}
\end{figure*}

As shown in Fig.~\ref{fig:framework}, we propose a universal framework for RS MTL based on VLM. The model follows a popular VLM paradigm. It uses a vision encoder and text tokenizer to process images and text inputs, respectively. The unified decoder based on a language model then process the bi-modal features and perform various tasks, such as RS image scene classification, question answering, captioning, grounding, and object detection.

Specifically, we develop a data curation engine consisting of data acquisition, offline processing, and online loading, which provides diverse images with textual prompts and golden responses for model training. To enable the model supporting images of arbitrary sizes, we design the dynamic resolution strategy, which handles input images of small, regular and UHR sizes respectively. The proposed Zoom-in Chain is designed to further enhance reasoning on UHR RS images. The final model can perform multiple tasks simultaneously. With the proposed auto-regressive aerial detection method, the model can perform the challenging aerial detection.

For the language branch, the input text is tokenized into a sequence of indices, where each index $i$ corresponds to a learnable embedding $\mathbf{t}_{i} \in \mathbb{R}^{D}$. The output sequence is then de-tokenized to produce the final textual response.

For the vision branch, a RS image is preprocessed (e.g., resized or dynamically rescaled) and encoded by a vision Transformer to obtain features $\mathbf{F} \in \mathbb{R}^{N_I \times D_I}$, where $N_I$ and $D_I$ denote the feature number and dimension. The prompt text is tokenized into $N_t$ embeddings $\mathbf{T}_t \in \mathbb{R}^{N_t \times D}$.
A bi-modal projection aligns visual embeddings with the language token space, generating $N_v$ visual tokens $\mathbf{T}_v \in \mathbb{R}^{N_v \times D}$, with $N_v \propto N_I$. The language model input is:
\begin{equation}
\mathbf{T} = \texttt{concat}(\mathbf{T}_v, \mathbf{T}_t) \in \mathbb{R}^{(N_v + N_t) \times D},
\end{equation}
where $\texttt{concat}(\cdot,\cdot)$ denotes token-wise concatenation.

During training, parameters $\theta$ are optimized via next-token prediction using cross-entropy loss:
\begin{equation}
\mathcal{L} = -\sum_{j=1}^{|\mathbf{r}|} P_j(\mathbf{r}, \mathbf{T}), \quad P_j(\mathbf{r}, \mathbf{T}) = \log P_{\theta}(\mathbf{r}j \mid \mathbf{r}{<j}, \mathbf{T}),
\end{equation}
where $\mathbf{r}=(\mathbf{r}_1,\dots,\mathbf{r}_T)$ is the response token sequence.

During inference, the model generates tokens auto-regressively until an end-of-sequence token is reached:
\begin{equation}
\mathbf{r}_j = \arg\max P_j(\mathbf{r}, \mathbf{T}) ~~ \text{or} ~~ \mathbf{r}_j \sim P_j(\mathbf{r}, \mathbf{T}),
\end{equation}
where the first denotes deterministic decoding (e.g., greedy or beam search) and the second stochastic sampling (e.g., top-$k$, nucleus sampling).

\subsection{Data Curation Engine}

\label{sec:data}

In contrast to conventional approaches that mainly conduct standardized evaluations on a single benchmark, this section highlights the crucial role of data curation in developing RS MTL models. 
Given the diversity and complexity of RS data—characterized by heterogeneous formats, noisy annotations, and inconsistent bounding box definitions. Hence, a well-curated data recipe is indispensable. 
To serve as a robust foundation for training RS multi-task VLMs, we design a data curation Engine, which is not a fixed dataset but a comprehensive and sustainable data framework encompassing the following three main parts.

\subsubsection{Data Acquisition}

In this work, the dataset was curated through three sequential stages. Initially, we collected data by following the data recipes of several representative open-source vision–language models. Specifically, we adopted the description–related subsets from the instruction tuning data of VHM~\cite{vhm} and GeoChat~\cite{geochat}, which cover tasks such as image classification, captioning, and visual question answering. These tasks can be further decomposed into subtasks, including modality recognition and resolution estimation. The refGeo~\cite{geoground} dataset was employed as the main grounding data source, while temporal multi-image data were drawn from TEOChatlas~\cite{irvin2024teochat}. To prevent degradation of the model’s general reasoning ability during continued training, we also incorporated a subset of general-purpose data sampled from LLaVA-OneVision’s recipe~\cite{li2025llavaonevision}, including chart interpretation, optical character recognition, and so on. By following these open-source data recipes, we indirectly surveyed and integrated diverse data sources.

Subsequently, we analyzed the limitations of the collected data and expanded the dataset using task-specific training set. We observed that existing RS VLMs rarely address object detection, which is crucial for fine-grained perception in RS. To fill this gap, we incorporated the DOTA-v1.0 dataset~\cite{dota}, thereby enriching the model’s detection-related learning capabilities.

Finally, for abilities that could not be obtained from open datasets, we constructed a synthetic data pipeline to generate new annotations. To enable the model’s zoom-in chain capability, we curated large-scale RS images and synthesized image–region–question triples. The detailed construction process is described in Section~\ref{zoomin}.

\subsubsection{Data Processing and Integrating}

Due to the diverse formats and task requirements of the collected datasets, as well as potential systematic noise, we performed additional offline preprocessing to integrate all data into our training framework.

We first removed all task descriptors, such as the ``[grounding]'', ``[refer]'', and ``[identify]'' tags used in previous works~\cite{geochat,vhm}. These descriptors tag the specific tasks. However, in open-world scenarios or novel tasks, instructions are typically expressed in natural language rather than through fixed descriptor tokens. Therefore, we replaced these descriptors with natural language prompts to better align with the real-world usage.

Next, we examined all bounding boxes in the datasets and categorized them into horizontal boxes, oriented boxes, and quadrilateral boxes. Their representations were then unified through consistent normalization and ordering to avoid any information mismatch. Corresponding prompts were designed for each box type. By default, horizontal boxes were used in grounding tasks, while quadrilateral boxes were adopted for detection tasks.

A unified data format was further established to standardize the integration. Conversational data followed the messages structure defined by OpenAI, object detection data were formatted according to the COCO convention, and grounding data adhered to the refGeo~\cite{geoground} schema.

Finally, we performed rule-based cleaning on systematic irregularities, such as removing redundant punctuation and spaces, and correcting typographical errors. For the Zoom-in Chain dataset, we applied tool-call formatting. The evaluation set was also processed in a similar manner to ensure consistency with the training data.

\subsubsection{Data Loading and Weighting}

After integration, the curated dataset was organized into multiple subset units. During training, we applied online preprocessing and dynamically controlled the sampling ratio of each subset. Consequently, the model was trained in a flexible and adaptive multi-task environment rather than on a fixed, pre-defined dataset.

We argue that the model should not rely solely on predefined prompts from the training stage. To enhance robustness, multiple agent prompts were designed for certain tasks, and one was randomly selected during training. For grounding and detection data, a unified formatting scheme was adopted. We also incorporated the JSON-based output format used in Qwen2.5-VL~\cite{qwen25vl}, accompanied by specific prompts, and randomly switched between standard and JSON outputs during training. In addition, a synonym replacement module was implemented to randomly substitute words with their synonyms, improving the model’s linguistic generalization. Standard data augmentation techniques, such as random resizing, were also applied to enhance multi-scale learning.

Each subset unit was assigned a sampling weight to guide data selection during training, analogous to controlling the flow rate of different ingredients in an automatic beverage dispenser. The sampling ratio is critical for multi-task learning: increasing the weight of more challenging tasks facilitates deeper learning, while adjusting the others helps mitigate catastrophic forgetting. In exploring optimal weighting strategies, we first conducted experiments with uniform ratios. Then, we increased the sampling proportion of tasks that underperformed relative to expectations. Finally, once all tasks reached or exceeded satisfactory performance, we fine-tuned the weights to achieve the best overall multi-task balance.

\subsection{Dynamic Resolution Strategy}

\label{sec:resolution}

Most existing RS VLMs (such as GeoChat~\cite{geochat}, VHM~\cite{vhm}, and GeoGround~\cite{geoground}) have the only fixed square input shape (such as $336 \times 336$ or $504 \times 504$). For each input image, they first pad the image to a square with zeros on the right or bottom, and then resize it to the input shape. Additionally, LRS-VQA~\cite{lrsvqa} and GeoLLaVA-8k~\cite{geollava8k} scale the input size to $2k\times2k$ and $8k\times8k$, respectively. They first cut a UHR RS image into slices of the fixed size and encode them into visual tokens. Then they prune the tokens to an amount comparable to the normal cases. In total, they pre-process the images only on a fixed shape or a small set of image shapes.

The proposed dynamic resolution strategy involves three interconnected aspects: supporting full-size input processing, scaling coordinate precision with input resolution, and curating training data to enhance learning across multiple resolutions. 

\subsubsection{Full-scale Visual Input}

\begin{figure}[tb]
  \centering
  \includegraphics[width=0.80\linewidth]{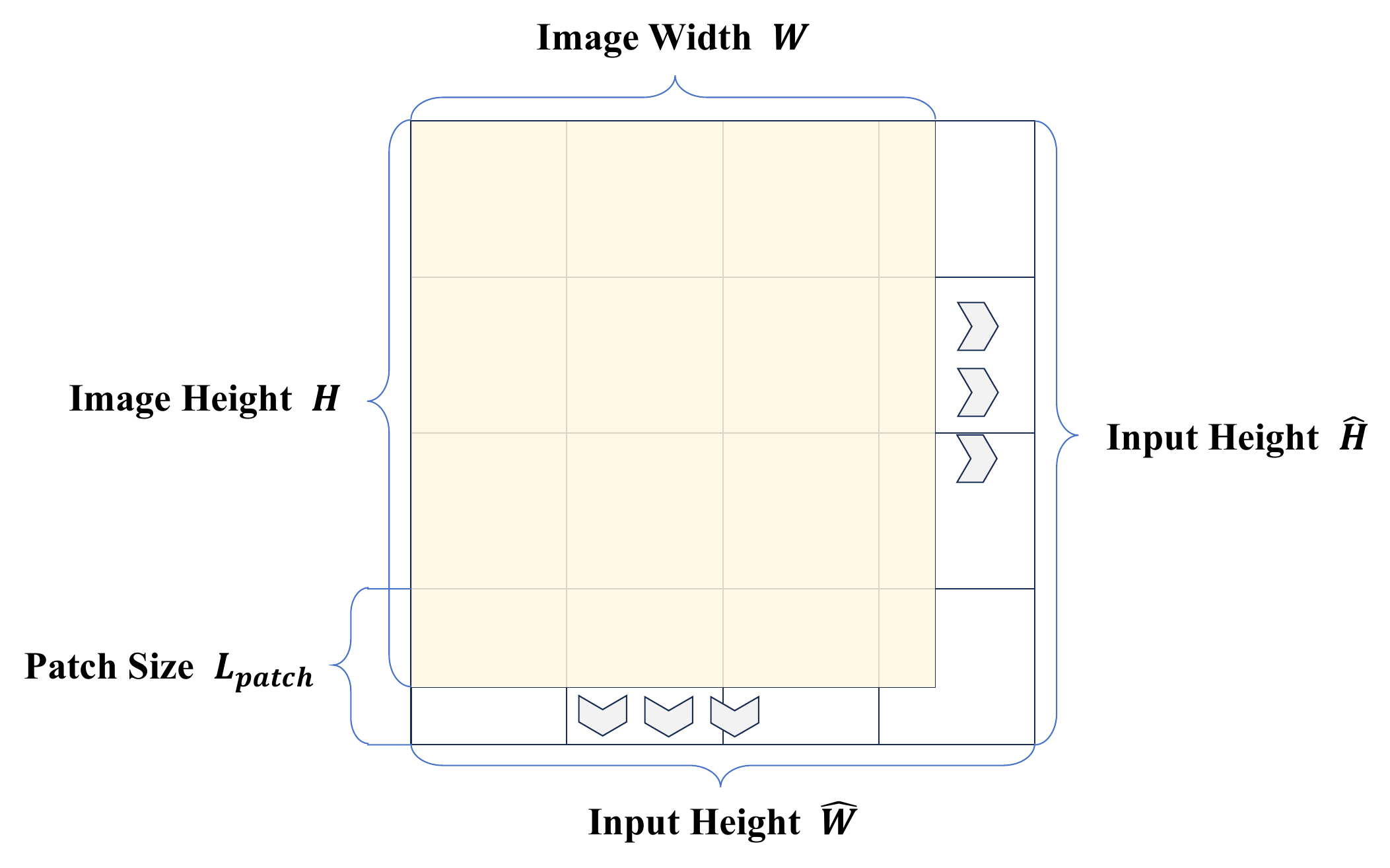}
  \vspace{-2mm}
  \caption{Schematic diagram of the native resolution input.}
  \vspace{-5mm}
  \label{fig:smart_resize}
\end{figure}

The native resolution scheme in Qwen2-VL~\cite{qwen2vl} inspire us to advance RS VLMs to accept inputs of arbitrary shapes. As shown in Fig.~\ref{fig:smart_resize}, let $H$ and $W$ indicate the height and width of a given RS image. $L_{patch}$ is the patch length corresponding to each visual token from the vision encoder. They first calculate the tightest shape that can wrap the input image by
\begin{equation}
    \small
    ( \hat{H}, \hat{W} ) = ( \lceil H / L_{patch} \rceil \times L_{patch}, \lceil W / L_{patch} \rceil \times L_{patch} ),
\end{equation}
where the $\lceil \cdot \rceil$ means the ceiling function. Then, they resize the image to $ ( \hat{H}, \hat{W} ) $ so that it can be exactly processed by the visual patch embedding.

This strategy allows the model to ingest images of arbitrary input sizes, which is well-suited to the diverse RS data. However, we should still set a range with a minimum scale to ensure adequate visual signal and a maximum scale due to constrained training resources. We divide the image sizes with the two bounds into three parts: small, regular, and UHR large. The small images are enlarged to ensure that there are enough visual tokens for the decoder to understand. For the UHR image, we design a zoom-in chain, which is introduced in \ref{zoomin}.

\subsubsection{Scalable Bounding Boxes}

For grounded or detection VLMs, spatial localization is achieved by directly generating numerical coordinates within textual outputs, which are extracted through regular expressions during inference.

However, existing RS VLMs often suffer from a mismatch between the coordinate resolution and the input image resolution. For instance, GeoChat~\cite{geochat} processes images at a fixed resolution of 504$\times$504, but its coordinate resolution is only 100$\times$100, leading to a fivefold loss in localization precision and poor performance on small objects. Conversely, GeoGround~\cite{geoground} employs a 336$\times$336 input resolution but defines coordinates at a much higher 1000$\times$1000 scale, resulting in more than half of the coordinate space being unused and excessive localization precision.

In this work, we adopt scalable bounding boxes, whose coordinate resolution dynamically aligns with the input image resolution, thereby avoiding both under- and over-precision issues. This design naturally adapts to varying input sizes and allows flexible control of inference cost depending on the required localization accuracy.

\subsubsection{Random Resizing}

To ensure robust performance across varying input image sizes, we applied dynamic scale augmentation during training. For each task, input images were randomly rescaled to different resolutions. In grounding and detection tasks, the corresponding bounding boxes were synchronously scaled to maintain spatial consistency. We observed that this scale-based augmentation significantly improved the model’s robustness to input-size variation. Moreover, the model trained under such conditions exhibited enhanced performance when performing high-resolution inference. This also enables a practical inference-time strategy, allowing users to adjust image resolution according to task requirements and computational constraints.

\subsection{Zoom-in Chain for UHR RS Images}

\begin{figure*}[t]
    \centering
    \includegraphics[width=\linewidth]{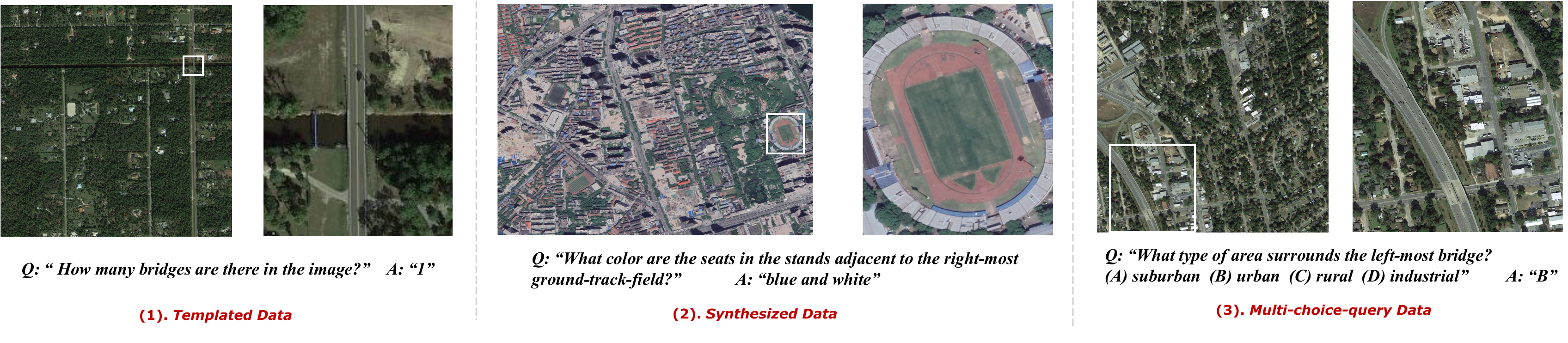}
    \vspace{-8mm}
    \caption{Examples of the three types of annotated data in the proposed LRS-VQA-Zoom.}
    \vspace{-5mm}
    \label{fig:lrsvqazoom}
\end{figure*}

\label{zoomin}

Previous works on understanding UHR RS images, such as LRS-VQA~\cite{lrsvqa} and GeoLLaVA-8k~\cite{geollava8k}, primarily focus on addressing the issue of excessive image tokens through visual token pruning. Although this approach has proven effective and computationally efficient, it typically requires additional training and is not well-suited for joint training with tasks using standard image resolutions.

We observed that when humans analyze UHR RS images on electronic devices, their workflow typically involves first scanning the entire image to identify regions of interest, then zooming into these regions before performing the actual task. Inspired by this workflow, we designed the Zoom-in Chain strategy for RS VLMs, as illustrated below:
\begin{align}
& \texttt{User}:~ \text{\textless Prompt\textgreater}+ \Imat_{\texttt{q}} + \text{\textless Question\textgreater} \nonumber \\
& ~~~~~~~~~~\texttt{Assistant}:~ \textcolor{blue}{[x1, y1, x2, y2]} \nonumber \\
& \texttt{User}:~ \texttt{Zoom-in}(\Imat_{\texttt{q}}, [x1, y1, x2, y2]) \nonumber \\
& ~~~~~~~~~~\texttt{Assistant}:~ \textcolor{blue}{\text{\textless Ground Truth\textgreater}}
\end{align}
The blue portions indicate the training labels, while the others are ignored for loss. Specifically, given a UHR RS image, we first downsample the image for initial processing. The model is prompted with instructions to predict the RoI, which is then cropped and fed into the model in native resolution. The final answer is obtained from both the initial and the new inputs, effectively mimicking the human zoom-in workflow for improved localization and task performance.

To enable the model to learn zoom-in capabilities during training, we construct a specialized instruction tuning dataset for UHR RS image perception, named LRS-VQA-Zoom. The data pipeline is initiated by collecting three public, large-scale UHR RS datasets: DOTA1.0~\cite{dota}, GLH-Bridge~\cite{li2024learning}, and STAR~\cite{li2024scene}. 

The methodology for generating the LRS-VQA-Zoom is extended from the pipeline in LRS-VQA~\cite{lrsvqa}. The final training corpus, totaling 302k samples, compromise three distinct subsets: 60k open-ended samples generated via rule-based templates, 159k open-ended samples synthesized using GPT-4V, and 83k samples in multi-choice-query form. Fig.~\ref{fig:lrsvqazoom} exhibits the examples from each subsets.

\subsubsection{Template-Generated Data (60k)}
This subset focuses on two open-ended question categories: counting and comparison.
For the counting data, the UHR image is first divided into a $3 \times 3$ grid (nine regions). Depending on the density of the target category, questions are formulated to query either the total count across the entire image or the count within a specific region. For the comparison data, these tasks involve comparing the relative quantities of two different object categories. For all samples in this subset, the absolute coordinates of the corresponding bounding boxes are preserved in the training data.

\subsubsection{GPT-4V-Synthesized Data (159k)}
This subset is designed to introduce greater question diversity.
First, we filter the original object detection labels to identify unique target instances, which serve as ``unique references". Subsequently, the ``coarse region" around each unique reference is cropped by applying a predefined padding margin. The dimensions of these coarse regions are suitable for processing by the GPT-4V model. We then prompt GPT-4V to generate diverse question-answer pairs based on these cropped regions. This process yields a rich variety of question types, including queries related to color, category, shape, status, spatial reasoning, and scene context (e.g., rural/urban). In this part of data, the coordinates of the horizontal bounding box defining the coarse region are recorded.

\subsubsection{Multi-choice-query Data (83k)}

To enhance the model's proficiency with mainstream evaluation formats (i.e., multiple-choice query (MCQ)) and to further diversify the training data, we converted a subset of 83k open-ended question answering samples into an MCQ format using an automated pipeline centered around the large language models. For each question-answer pair, excluding simple binary (yes/no) queries, we prompted the GPT-4 to generate three plausible but incorrect ``distractors" and return them alongside the original correct answer in a structured JSON format. This output was then systematically validated to ensure it contained four unique options. Finally, to prevent positional bias, the options were randomly shuffled, and the sample was formatted to include the question, four choices prefixed with letters (A, B, C, D), and the letter corresponding to the ground truth answer.

\subsection{Auto-regressive Aerial Detection}

\label{sec:detection}

In this paper, we investigate multi-class oriented aerial object detection. To enable the RS VLM to perform dense detection in aerial images, we propose an auto-regressive detection paradigm, representing detection outputs directly in textual form, as illustrated in the right part of Fig.~\ref{fig:framework}. Specifically, we propose a normalization procedure for model responses and a novel evaluation metric to facilitate fair comparisons between the RS VLM detectors and conventional detectors.

\subsubsection{Response Normalization}

\label{subsec:detection-method}

In the aerial object detection task, each object is represented by its class label and an 8-parameter quadrilateral bounding box 
$\mathbf{o}=(n_\mathbf{o}, x_{1\mathbf{o}}, y_{1\mathbf{o}}, x_{2\mathbf{o}}, y_{2\mathbf{o}}, x_{3\mathbf{o}}, y_{3\mathbf{o}}, x_{4\mathbf{o}}, y_{4\mathbf{o}})$, 
where $(x_{i\mathbf{o}}, y_{i\mathbf{o}})$ denote the coordinates of the polygon vertices in clockwise order. 
The vertex with the smallest vertical coordinate is designated as the starting point. The class label $n_{\mathbf{o}}$ corresponds to one of the $c$ predefined categories $\{C_1, C_2, ..., C_c\}$.

To standardize detection annotations, a consistent template is employed to ensure both uniqueness and order. 
For each input image, the model outputs detected objects in a structured sequence. 
Specifically, detection results are first grouped by category and sorted alphabetically by category name. 
Within each category, the bounding boxes are further ordered according to the position of their designated starting vertex.

During our extension of LMMRotate~\cite{lmmrotate}, we observed a subtle yet important issue. In the LMMRotate, images without any objects were removed from the training set to improve efficiency, following common practices in conventional aerial detector. However, this approach can be detrimental when training a VLM, as encountering object-free images during inference often leads the model to hallucinate, producing false positive detections. 
To address this, RSCoVLM retains images without objects in the training process and explicitly trains the model to output ``There is none.'' for such cases, thereby mitigating hallucinations and improving detection reliability.

Our VLM is capable of detecting multiple object categories within an aerial image, with both category labels and bounding box coordinates included in its output. 
During inference, detection results can be retrieved directly from the model response using straightforward regular expression parsing. 
Furthermore, unlike most traditional detectors that require post-processing procedures such as non-maximum suppression (NMS) to address overlapping or redundant detections, the VLM inherently avoids these issues.

\subsubsection{Evaluation Metrics}

In conventional aerial detection tasks, mean average precision (mAP) is widely employed as the evaluation metric, requiring bounding boxes, class labels, and confidence scores for all detected objects. However, as discussed earlier, our model responses only include object categories and their corresponding spatial coordinates, which implies that vision-language models based on the proposed detection approach cannot directly produce mAP results.

In mAP calculation, confidence scores are used to rank detector predictions, which directly influences the accumulation of true and false positives along the PR curve. The conventional object detection models often retain low-confidence yet actually false-positive predictions to maintain higher mAP scores. For instance, several DETR-based models evaluate all 900 proposals per image even when only a few objects are present. Since vision-language models typically yield only a limited number of confident predictions, we are constrained to assign fixed or randomized confidence scores to enable mAP computation. However, the lack of well-calibrated confidence estimates still places vision-language models at a substantial disadvantage in mAP-based evaluation with the adaptation, even when visual inspection suggests that VLMs achieve performance comparable to conventional detectors.

In practice, an object detection prediction is composed of a location and a category. The confidence scores are primarily used to filter out low-confidence detections via thresholding, as part of the detection result visualization procedure. If we first filter out low-confidence predictions using a threshold and then randomize the confidence scores, we could remove the influence of confidence on the mAP calculation. We denoted this as mAP with no conference scores, i.e., $\text{mAP}_{\text{nc}}$.

\begin{figure}[tb]
  \centering
  \includegraphics[width=\linewidth]{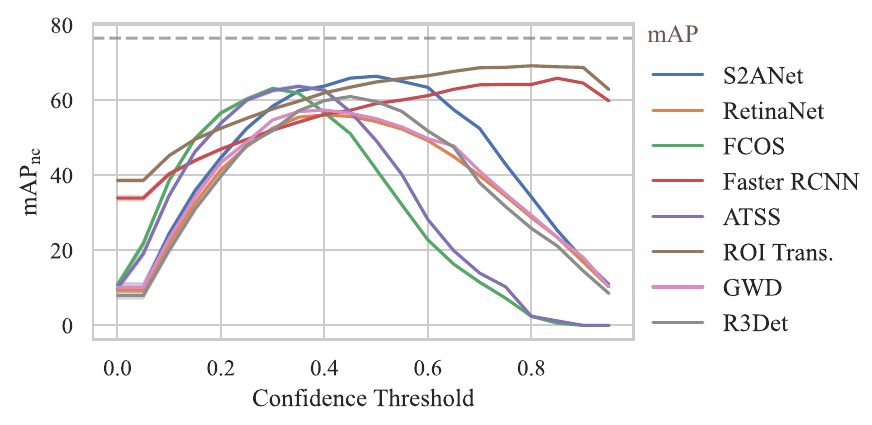}
  \vspace{-3mm}
  \caption{The impact of confidence scores on $\text{mAP}_{\text{nc}}$ with error bands. The colored lines record the variation trends of $\text{mAP}_{\text{nc}}$ for the popular conventional detector on DOTA-v1.0~\cite{dota} (trained and evaluated on both the `train` split and the `validation` split) dataset under different confidence thresholds.}
  \vspace{-5mm}
  \label{fig:score_impact}
\end{figure}

Figure~\ref{fig:score_impact} illustrates the $\text{mAP}_{\text{nc}}$ of several detectors. The horizontal axis illustrates the variation of $\text{mAP}_{\text{nc}}$ with increasing filtering thresholds. For each detector, the validation results were computed over ten runs with different random seeds, and the solid line represents the mean of these ten runs. The solid line in the figure is enveloped by a light-colored error band to indicate the effect of randomness on the results. Initially, as the filtering threshold increases, the detector's performance improves because low-confidence false positives are progressively filtered out. After reaching a peak, overly strict filtering begins to remove true positives as well, causing the results to decline. Notably, this peak is still lower than the mAP metric that incorporates confidence scores. Furthermore, the error band is barely visible without magnification, indicating that although randomness is involved, the variance of $\text{mAP}_{\text{nc}}$ is very small.

Instead of introducing an additional mechanism to estimate confidence for VLM-based detectors, we argue that confidence should not be a prerequisite when evaluating or comparing detection performance between VLMs and conventional detectors. Detection annotations and outputs inherently consist of class labels and bounding boxes, while confidence scores are auxiliary byproducts generated during inference. They may facilitate postprocessing but are not indispensable for evaluating model accuracy. Therefore, we advocate employing confidence-independent metrics such as mean F1-score ($\text{mF}_1$) and $\text{mAP}_{\text{nc}}$ for a more equitable evaluation. {Additionally, the small variance of $\text{mAP}_{\text{nc}}$ also demonstrates its stability as a metric.}

Finally, for benchmarks such as DOTA~\cite{dota} and FAIR1M~\cite{FAIR1M}, where public test sets are unavailable and online evaluation servers rely solely on mAP, we recommend adopting $\text{mAP}_{\text{nc}}$ as the primary evaluation metric to ensure consistent and fair assessment across different model types.

\section{Experiment}

\begin{figure*}[t]
    \centering
    \includegraphics[width=\linewidth]{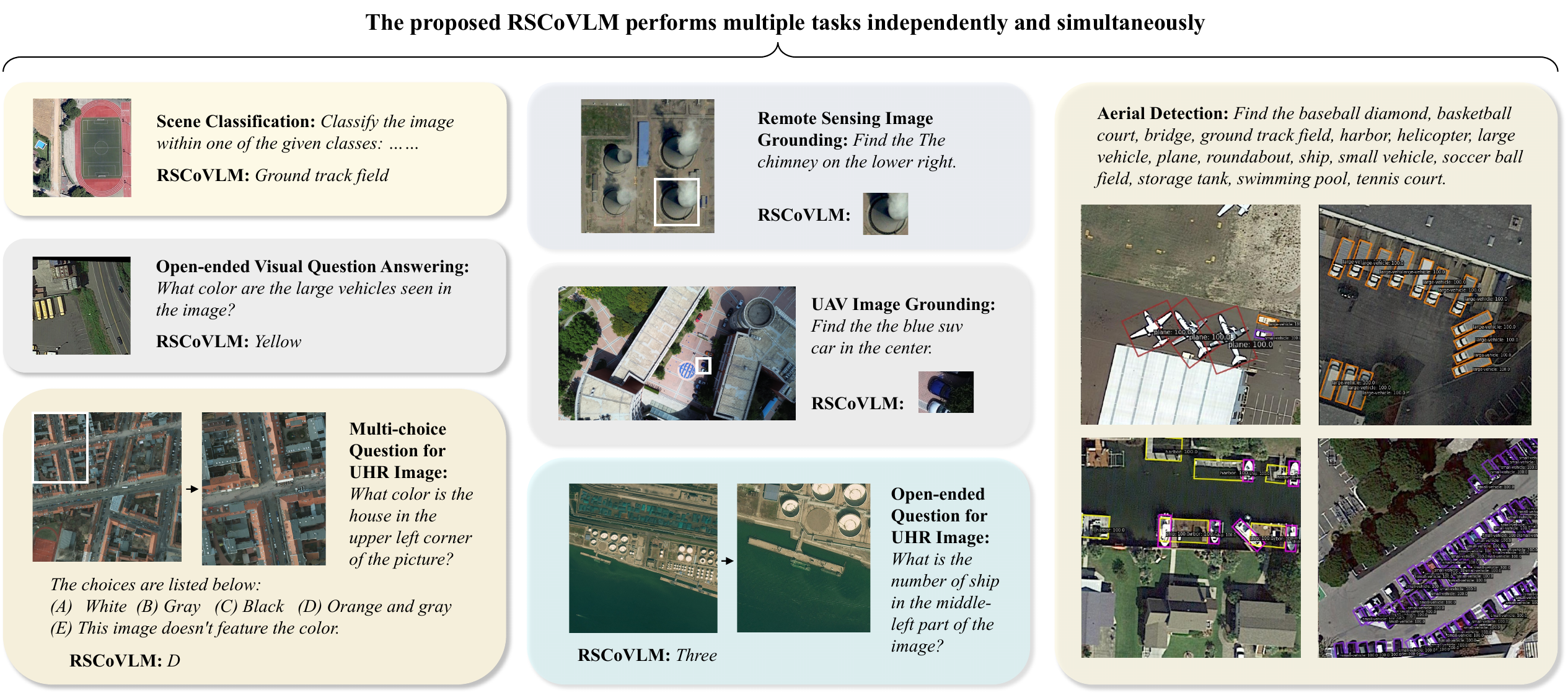}
    \vspace{-2mm}
    \caption{Demonstration of RSCoVLM’s capabilities on several commonly used tasks, including scene classification, open-ended and multiple-choice question answering for regular and UHR images, visual grounding in aerial and UAV images, and aerial object detection. In particular, the visualized results of aerial detection are especially impressive.}
    \vspace{-5mm}
    \label{fig:result}
\end{figure*}

\begin{table*}[tb]
\renewcommand{\arraystretch}{1.3}  %
\centering
\caption{Comparison Results of State-of-the-art Vision-Language Models and Our Model on the LRS-VQA Benchmark}
\label{tab:res-lrs-vqa}
\begin{tabular}{l|cc|ccc|c}
\hline
\textbf{Method}                                & \textbf{Model Size}          & \textbf{Max pixels}            & \textbf{LRS-FAIR} & \textbf{LRS-Bridge} & \textbf{LRS-STAR} & \textbf{Avg. Acc} \\ \hline
LLaVA-1.5~\cite{llava15}                       & 7B                           & 0.1M                           & 18.76             & 30.70               & 22.63             & 24.03             \\ \hline
LLaVA-UHD-v2~\cite{llavauhdv2}                 & 7B                           & 0.7M                           & 22.82             & 32.57               & 26.08             & 27.16             \\ \hline
Qwen2-VL~\cite{qwen2vl}                        & 7B                           & 11.1M                          & 23.80             & 38.12               & 27.87             & 29.93             \\ \hline
Qwen2.5-VL~\cite{qwen25vl}                     & 7B                           & 12.8M                          & 19.66             & 35.82               & 26.70             & 27.39             \\ \hline
\multirow{2}{*}{\makecell{Qwen3-VL~\cite{qwen25vl}}}      & 8B                           & 16.8M                          & \textbf{27.98}    & \textbf{38.56}      & \textbf{32.04}    & \textbf{32.86}    \\ \cline{2-7} 
                                               & A3B-30B                      & 16.8M                          & \textbf{27.63}    & 38.81               & 30.54             & 32.33             \\ \hline
InternVL2.5-MPO~\cite{mpo}                 & 8B                           & 2.4M                           & 24.95             & 34.59               & 25.14             & 28.23             \\ \hline
InternVL3~\cite{internvl3}                     & 8B                           & 2.4M                           & 22.49             & 38.09               & 26.36             & 28.98             \\ \hline
\multirow{2}{*}{\makecell{InternVL3.5~\cite{internvl35}}} & 8B                           & 2.4M                           & 25.14             & 35.50               & 26.86             & 29.17             \\ \cline{2-7} 
                                               & A3B-30B                      & 2.4M                           & 16.83             & 37.05               & 22.15             & 25.34             \\ \hline
Mimo-VL~\cite{mimovl}                   & 7B                           & 12.8M                          & 16.51             & 20.04               & 27.11             & 21.22             \\ \hline
GeoChat~\cite{geochat}                        & 7B                           & 0.3M                           & 20.18             & 24.54               & 13.75             & 19.49             \\ \hline
LLaVA-1.5~\textcolor{gray}{\emph{+ SFT. on LRS-VQA}}~\cite{lrsvqa}                & 7B                           & 0.1M                           & 22.97             & 36.89               & 27.48             & 29.11             \\ \hline
LLaVA-Next~\textcolor{gray}{\emph{+ SFT. on LRS-VQA}}~\cite{lrsvqa}               & 7B                           & 2.8M                           & 21.85             & 38.24               & 26.67             & 28.92             \\ \hline
\textbf{RSCoVLM}                               & \multirow{2}{*}{\makecell{\textbf{7B}}} & \multirow{2}{*}{\makecell{\textbf{1.0M}}} & 27.37             & \textbf{42.42}      & \textbf{31.77}    & \textbf{33.85}    \\ \cline{1-1} \cline{4-7} 
~~\textcolor{gray}{\emph{+ Zoom-in Chain}}     &                              &                                & \textbf{42.42}    & \textbf{49.56}      & \textbf{45.15}    & \textbf{45.71}    \\ \hline
\end{tabular}
\end{table*}

\begin{table*}[tb]
\renewcommand{\arraystretch}{1.3}  %
\centering
\caption{Comparison Results of State-of-the-art Vision-Language Models and Our Model on Visual Grounding Benchmarks}
\label{tab:res-vg}
\begin{tabular}{l|c|cc|cc|c|c|c|c}
\hline
\multirow{2}{*}{\makecell{\textbf{Method}}}                                       & \multirow{2}{*}{\makecell{\textbf{Input Size}}} & \multicolumn{2}{c|}{\textbf{DIOR-RSVG}} & \multicolumn{2}{c|}{\textbf{RSVG}} & \multirow{2}{*}{\makecell{\textbf{GeoChat-VG}}} & \multirow{2}{*}{\makecell{\textbf{VRSBench-VG}}} & \multirow{2}{*}{\makecell{\textbf{AVVG}}} & \multirow{2}{*}{\makecell{\textbf{Avg. Acc.}}} \\ \cline{3-6}
                                                                       &                                      & \textbf{val}       & \textbf{test}      & \textbf{val}     & \textbf{test}   &                                        &                                         &                                &                                     \\ \hline
Qwen-VL-Chat~\cite{qwen2vl}                                             & $448\times448$                       & 32.01              & 32.22              & 4.66             & 2.04            & 35.36                                  & 31.07                                   & 0.31                           & 19.66                               \\
GeoChat~\cite{geochat}                                                 & $504\times504$                       & 23.35              & 24.05              & 3.08             & 2.04            & 22.74                                  & 11.52                                   & 0.28                           & 12.44                               \\
LHRS-Bot~\cite{lhrsbot}                                                & $224\times224$                       & 17.04              & 17.59     & 0.95    & 1.56   & 3.25                          & 1.19                           & 0.00                  & 5.94                       \\
VHM~\cite{vhm}                                                         & $336\times336$                       & \textbf{-}         & 48.04     & \textbf{-}       & \textbf{-}      & \textbf{-}                             & \textbf{-}                              & \textbf{-}                     & \textbf{-}                          \\
Qwen2.5-VL~\cite{qwen25vl}                                             & Dynamic                              & 43.64              & 45.26              & 19.73            & 21.27           & 42.99                                  & 44.50                                   & 7.64                           & 32.15                               \\
Qwen-VL~\textcolor{gray}{\emph{+ SFT. on refGeo}}~\cite{geoground}     & $448\times448$                       & 58.65              & 58.76              & 12.99            & 10.59           & 41.75                                  & 47.38                                   & 9.53                           & 34.24                               \\
GeoChat~\textcolor{gray}{\emph{+ SFT. on refGeo}}~\cite{geoground}     & $504\times504$                       & 60.27              & 61.96              & 16.32            & 14.67           & 56.99                                  & 51.36                                   & 11.52                          & 39.01                               \\
LLaVA-1.5-7B~\textcolor{gray}{\emph{+ SFT. on refGeo}}~\cite{geoground}& $336\times336$                       & 64.46              & 65.98              & 19.98            & 20.95  & 63.76                                  & 57.17                                   & 15.05                 & 43.91                               \\
GeoGround~\cite{geoground}                                             & $336\times336$                       & \textbf{77.18}     & \textbf{77.73}     & \textbf{27.64}   & \textbf{26.65}  & \textbf{70.24}                         & 66.04                          & \textbf{21.58}                 & \textbf{52.44}                      \\ \hline
\textbf{RSCoVLM}                                                       & Dynamic                              & \textbf{83.56}     & \textbf{84.55}     & \textbf{54.04}   & \textbf{53.79}  & \textbf{76.39}                         & \textbf{79.73}                          & \textbf{29.40}                 & \textbf{65.92}                      \\
~~\textcolor{gray}{\emph{+ Min Size}}                                  & $224\times224$                       & 66.56              & 67.64              & 21.23            & 20.70           & 21.43                                  & \textbf{67.50}                          & 0.85                           & 37.99                               \\
~~\textcolor{gray}{\emph{+ Small Size}}                                & $336\times336$                       & \textbf{75.22}     & \textbf{75.86}     & \textbf{34.72}   & \textbf{35.79}  & \textbf{70.17}                         & \textbf{75.79}                          & \textbf{25.10}                 & \textbf{56.09}                      \\ \hline
\end{tabular}
\end{table*}

In this section, the RSCoVLM is evaluated on benchmarks across various tasks, demonstrating the promising multi-task capabilities. We firstly provide detailed implementation specifications to facilitate reproducibility. Then, we compare our model with state-of-the-art methods on various RS understanding and perception tasks with different input resolutions. 

Fig.~\ref{fig:result} presents the demonstration of RSCoVLM’s capabilities on several commonly used tasks. Notably, all tasks are accomplished using a single RSCoVLM model, demonstrating its impressive multi-task capability.

\subsection{Reproducibility Details}

We use \href{https://huggingface.co/Qwen/Qwen2.5-VL-7B-Instruct}{Qwen2.5-VL-7B-Instruct}~\cite{qwen25vl} as the foundation model of RSCoVLM. The model is optimized with AdamW, employing a weight decay of 0.1. We train the full model with a base learning rate of 2$\times$10$^{-\text{6}}$, following a cosine learning rate schedule with a linear warmup over the first 5\% of training steps. The total batch size is set to 32, and the maximum sequence length is 6,144 tokens. The input images are constrained to resolutions between 224$\times$224 and 1,008$\times$1,008 pixels.

We have released the codebase on the \href{\githuburl}{\faGithub~GitHub repository} and uploaded the whole well-collected data folder and model weights to the \href{\huggingfaceurl}{\hflogo~HuggingFace repository}. The codebase is implemented concisely, leveraging resource-efficient and effective training techniques. To save GPU memory, we adopt DeepSpeed-ZeRO-Stage-1~\cite{zero} and gradient checkpointing. For improved computational efficiency, we utilize BFloat16 precision and Flash-Attention-2~\cite{dao2024flashattention} during both training and evaluation. Additionally, Liger Kernel~\cite{hsu2025ligerkernel} is employed to accelerate training, and vLLM~\cite{vllm} is used for faster inferencing. All experiments are conducted on \href{https://www.volcengine.com/}{VolcEngine} high-performance computing clusters equipped with NVIDIA A800 GPUs. We'll maintain the repositories and update the latest code, model and data in our future research progress.

\subsection{Evaluation on Large RS Imagery}

\subsubsection{Benchmark and Metric}

The LRS-VQA~\cite{lrsvqa} is the latest visual question answering benchmark for large RS images. It features 7,333 question-answer pairs across 8 categories, including count, color, category, shape, status, reasoning, rural/urban classification, and target background. The images in this benchmark reach up to 27,328 pixels in length and have an average size of 7,099$\times$6,329 pixels. 

There are three subsets, corresponding to three data sources: FAIR1M~\cite{FAIR1M}, GLH-Bridge~\cite{li2024learning}, and STAR~\cite{li2024scene}. The official scoring implementation first calculates accuracy for each source and task, and then computes average accuracy (AA) across tasks for each sources. The AAs for each subset are reported.

\subsubsection{Results}

The results are presented in Table~\ref{tab:res-lrs-vqa}. The max pixels numbers of each models are also provided. It can be seen that the average pixel number of LRS-VQA (about 45 million) has been larger than the largest pixels uplimit (16.8 million for Qwen3-VL~\cite{qwen25vl}). 

As shown in the table, the proposed Zoom-in Chain approach substantially enhances the model’s performance, achieving an overall improvement of 35\% compared to the baseline that inference solely. 

Furthermore, our model demonstrates stronger foundational capabilities than other competing models, approaching the performance of the leading Qwen3-VL-8B~\cite{qwen25vl}, while utilizing a slightly smaller parameter count and a significantly lower maximum input resolution. Our model also outperforms other RS foundation models, including GeoChat and the officially fine-tuned LLaVA-Next model for LRS-VQA~\cite{lrsvqa}.

\subsection{Evaluation on Visual Grounding}

\subsubsection{Benchmark and Metric}

We follow GeoGround~\cite{geoground} for visual grounding evaluation because of its strong emphasis on comprehensiveness, fairness, and transparency. The evaluation incorporates the validation and test sets of DIOR-RSVG and RSVG~\cite{rsvg}, the visual grounding portions of GeoChat-Bench~\cite{geochat} and VRSBench~\cite{vrsbench}, as well as AVVG benchmark~\cite{geoground} for images captured by unmanned aerial vehicle. The evaluation details have strictly aligned with GeoGround. We directly adopted the splits and annotations provided by GeoGround for all benchmarks~\cite{geoground}.

We follow common practice to utilize Acc@0.5 as the evaluation metric, which regards the predicting that has an Intersection over Union (IoU) greater than 0.5 with the ground truth as a successful localing.

\subsubsection{Results}

Table~\ref{tab:res-vg} presents the results, along with the corresponding input sizes for each model. The input resolutions of existing RS VLMs, including GeoChat~\cite{geochat}, LHRS-Bot~\cite{lhrsbot}, VHM~\cite{vhm}, and GeoGround~\cite{geoground}, are fixed and typically smaller than 512$\times$512. In contrast, only general-purpose VLMs such as Qwen2.5-VL~\cite{qwen25vl} and MiMo-VL~\cite{mimovl} support dynamic input resolution, enabling flexible adaptation to varying input sizes.

Our model demonstrates substantially superior performance across all benchmarks. It surpasses the previously best-performing visual-language model specialized for RS grounding, GeoGround, by approximately 25.7\%, and outperforms all baselines that were supervised-finetuned on refGeo.

We further conducted experiments using fixed low-resolution inputs to intentionally weaken our model’s performance. Even at the minimal input size of 224$\times$224, our model maintains strong capability; however, such a small resolution severely limits image clarity, causing small objects to occupy only a few pixels and become indistinguishable. In particular, performance on AVVG drops sharply, indicating that a 224$\times$224 resolution is highly impractical for RS grounding. When evaluated at 336$\times$336, which aligns with the input size of other comparison methods, our model still achieves state-of-the-art results.

We attribute this performance advantage to three primary factors. First, the support for dynamic input resolution allows the model to perform inference at native resolution without downsampling, preserving visual detail. Second, the multi-resolution augmentation strategy employed during training enables the model to generalize effectively across diverse resolutions and computational budgets. Finally, auxiliary localization-related tasks, such as object detection and zoom-in refinement, further strengthen the model’s grounding ability and robustness.

\subsection{Evaluation on Object Detection}

\begin{table*}[tb]\Large
\renewcommand{\arraystretch}{1.5}  %
\centering
\caption{Comparison Results of State-of-the-art Aerial Detectors and Our Model on DOTA-v1.0 Benchmark}
\label{tab:dota}
\resizebox{\textwidth}{!}{%
\begin{tabular}{lc|ccccccccccccccc|c|c|c}
\hline
\multicolumn{1}{l|}{\textbf{Method}}               & \textbf{score} & \textbf{PL}    & \textbf{BD}    & \textbf{BR}    & \textbf{GTF}   & \textbf{SV}    & \textbf{LV}    & \textbf{SH}    & \textbf{TC}    & \textbf{BC}    & \textbf{ST}    & \textbf{SBF}   & \textbf{RA}    & \textbf{HA}    & \textbf{SP}    & \textbf{HC}    & \textbf{$\text{AP}_{\text{nc}50}$} & \textbf{$\text{AP}_{\text{nc}75}$} & \textbf{$\text{AP}_{\text{nc}50:95}$} \\ \hline
\multicolumn{1}{l|}{GWD~\cite{gwd}}                & 0.40           & 72.07          & 58.95          & 24.26          & 35.92          & 63.37          & 52.24          & 66.63          & 86.85          & 61.54          & 59.15          & 23.45          & 45.84          & 36.10          & 49.50          & 26.08          & 50.80                              & 28.67                              & 28.98                                 \\
\multicolumn{1}{l|}{R3Det~\cite{r3det}}            & 0.45           & 73.32          & 59.51          & 31.59          & 43.37          & \textbf{64.94} & 63.37          & \textbf{75.61} & \textbf{89.35} & 63.84          & 66.95          & \textbf{34.99} & 45.60          & 46.54          & 50.17          & 14.48          & 54.91                              & 29.21                              & 30.08                                 \\
\multicolumn{1}{l|}{ATSS~\cite{atss}}              & 0.35           & 72.98          & 60.67          & 25.82          & 42.91          & \textbf{65.23} & \textbf{65.32} & 75.22          & \textbf{89.78} & \textbf{71.61} & \textbf{70.12} & 28.04          & 43.19          & 47.79          & 58.28          & 28.60          & 56.37                              & \textbf{34.62}                     & \textbf{33.05}                        \\
\multicolumn{1}{l|}{Faster RCNN~\cite{fasterrcnn}} & 0.85           & 73.49          & \textbf{67.37} & \textbf{32.50} & 43.19          & 62.92          & 63.13          & 73.95          & 88.79          & \textbf{73.77} & 66.33          & 25.23          & \textbf{48.89} & \textbf{53.03} & 56.94          & 31.17          & \textbf{57.38}                     & \textbf{32.98}                     & \textbf{32.96}                        \\
\multicolumn{1}{l|}{FCOS~\cite{fcos}}              & 0.30           & 72.09          & 56.32          & 32.02          & 27.79          & 64.28          & 63.83          & \textbf{75.75} & 89.21          & 68.11          & \textbf{67.82} & 27.30          & 37.15          & 46.64          & \textbf{58.98} & 19.53          & 53.79                              & 32.13                              & 31.49                                 \\
\multicolumn{1}{l|}{CSL~\cite{csl}}                & 0.40           & 71.57          & 53.57          & 19.82          & 35.90          & 64.04          & 44.96          & 66.35          & 87.54          & 61.33          & 59.26          & \textbf{29.27} & 39.56          & 36.51          & 49.80          & 17.38          & 49.12                              & 29.03                              & 28.72                                 \\
\multicolumn{1}{l|}{S2A-Net~\cite{s2anet}}         & 0.50           & 72.96          & 61.96          & \textbf{36.00} & \textbf{45.99} & \textbf{66.24} & \textbf{65.61} & \textbf{77.08} & \textbf{89.34} & \textbf{73.27} & \textbf{69.25} & \textbf{31.78} & \textbf{46.64} & \textbf{55.02} & 52.02          & 35.40          & \textbf{58.58}                     & 29.53                              & 31.98                                 \\ \hline
\multicolumn{2}{l|}{\textbf{RSCoVLM}}                               & \textbf{77.15} & \textbf{64.86} & 23.90          & \textbf{45.34} & 44.87          & 38.96          & 57.64          & 87.22          & 57.73          & 49.42          & 23.31          & 51.87          & 37.01          & 54.92          & \textbf{54.91} & 51.27                              & 25.75                              & 27.60                                 \\
\multicolumn{2}{l|}{~~\textcolor{gray}{\emph{+ Max Mode}}}          & \textbf{73.95} & 63.01          & 27.84          & 40.41          & 56.86          & 55.37          & 71.00          & 89.12          & 61.69          & 64.95          & 19.54          & 41.91          & 44.21          & 55.01          & \textbf{52.27} & 54.48                              & 31.04                              & 31.38                                 \\
\multicolumn{2}{l|}{\textbf{RSCoVLM-det}}                           & \textbf{73.52} & \textbf{64.68} & 26.89          & \textbf{47.18} & 52.57          & 52.71          & 59.33          & 89.16          & 63.17          & 61.43          & 18.91          & 45.96          & 47.62          & \textbf{59.19} & \textbf{70.24} & 55.50                              & 30.78                              & 31.75                                 \\
\multicolumn{2}{l|}{~~\textcolor{gray}{\emph{+ Max Mode}}}          & 69.04          & 64.44          & \textbf{33.32} & 44.67          & 56.21          & \textbf{66.47} & 73.71          & 87.59          & 61.38          & 63.95          & 22.41          & \textbf{46.63} & \textbf{47.90} & \textbf{59.82} & 50.82          & \textbf{56.56}                     & \textbf{33.88}                     & \textbf{33.66}                        \\ \hline
\end{tabular}%
}
\end{table*}

\begin{table*}[!tb]
\renewcommand{\arraystretch}{1.3}  %
\centering
\caption{Comparison Results of State-of-the-art Vision-Language Models and Our Model on Five Scene Classification Benchmarks}
\label{tab:res-cls}
\begin{tabular}{l|c|ccccc}
\hline
\textbf{Method}                               & \textbf{Model Size} & \textbf{AID}                          & \textbf{UCMerced} & \textbf{METER-ML} & \textbf{NWPU-RESISC45} & \textbf{WHU-RS19} \\ \hline
MiniGPTv2~\cite{minigptv2}                    & 7B                  & \phantom{00.00}-~$|$~32.96\phantom{-} & -                 & 14.29             & 28.15                  & 64.80             \\ \hline
LLaVA-1.5~\cite{llava15}                      & 7B                  & \phantom{00.00}-~$|$~31.10\phantom{-} & -                 & 21.73             & 34.96                  & 54.55             \\ \hline
Qwen-VL-Chat~\cite{qwen2vl}                    & 7B                  & \phantom{00.00}-~$|$~55.30\phantom{-} & -                 & 38.77             & 42.73                  & 72.25             \\ \hline
Qwen2.5-VL~\cite{qwen25vl}                    & 7B                  & 63.63~$|$~62.73                       & 70.90             & 56.64             & 64.98                  & 76.20             \\ \hline
\multirow{2}{*}{Qwen3-VL~\cite{qwen25vl}}      & 8B                  & 70.84~$|$~66.67                       & 79.90             & 60.88             & 68.86                  & 87.80             \\ \cline{2-7} 
                                              & A3B-30B             & 71.75~$|$~68.87                       & 80.19             & 64.07              & 70.22                  & 87.70             \\ \hline
InternVL2.5-MPO~\cite{mpo}              & 8B                  & 69.38~$|$~64.23                       & 62.90             & 55.04             & 59.21                  & 80.20             \\ \hline
InternVL3~\cite{internvl3}                    & 8B                  & 67.78~$|$~63.40                       & 67.29             & 59.65             & 64.32                  & 86.40             \\ \hline
\multirow{2}{*}{InternVL3.5~\cite{internvl35}} & 8B                  & 77.03~$|$~75.00                       & 83.43             & 51.33             & 92.57                  & 91.70             \\ \cline{2-7} 
                                              & A3B-30B             & 82.45~$|$~79.17                       & \textbf{86.00}    & 46.19             & \textbf{98.38}         & \textbf{97.10}   \\ \hline
Mimo-VL~\cite{mimovl}                  & 7B                  & 66.13~$|$~67.20                       & 69.14             & 54.51             & 64.35                  & 86.10             \\ \hline
LHRSBot~\cite{lhrsbot}                        & 7B                  & \phantom{00.00}-~$|$~\textbf{91.26}\phantom{-} & -                 & 69.81             & 83.94                  & 93.17             \\ \hline
GeoChat~\cite{geochat}                        & 7B                  & \phantom{-}72.00~$|$~-\phantom{00.00} & 84.40             & -                 & -                      & -                 \\ \hline
TEOChat~\cite{irvin2024teochat}                        & 7B                  & \phantom{-}80.90~$|$~-\phantom{00.00} & \textbf{86.30}    & -                 & -                      & -                 \\ \hline
LHRS-Bot-Nova~\cite{lhrsbot-nova}             & 7B                  & \phantom{-}\textbf{83.06}~$|$~-\phantom{00.00} & -                 & \textbf{72.74}    & 83.97                  & \textbf{96.20}    \\ \hline
SkysenseGPT~\cite{luo2024skysensegpt}                & 7B                  & \phantom{-}\textbf{88.16}~$|$~-\phantom{00.00} & -                 & 40.00             & 90.06                  & 95.50             \\ \hline
VHM~\cite{vhm}                                & 7B                  & \phantom{00.00}-~$|$~\textbf{91.70}\phantom{-} & -                 & \textbf{72.74}    & \textbf{94.54}         & 95.80             \\ \hline
ScoreRS~\cite{scorers}                        & 7B                  & \phantom{00.00}-~$|$~85.90\phantom{-} & -                 & \textbf{74.42}    & 91.59                  & \textbf{96.30}    \\ \hline
\textbf{RSCoVLM}                              & \textbf{7B}         & \textbf{88.44~$|$~94.30}              & \textbf{94.52}    & \textbf{75.93}    & \textbf{98.25}         & 95.80             \\ \hline
\end{tabular}
\end{table*}

\begin{table*}[tb]
\renewcommand{\arraystretch}{1.3}  %
\centering
\caption{Comparison Results of State-of-the-art Vision-Language Models and Our Model on Two VQA Benchmarks}
\label{tab:res-vqa}
\begin{tabular}{l|cccccc|c}
\hline
\multirow{2}{*}{\textbf{Method}}  & \multicolumn{6}{c|}{\textbf{RSVQA Benchmark}}                                                                                         & \multirow{2}{*}{\textbf{\begin{tabular}[c]{@{}c@{}}VRSBench \\ VQA\end{tabular}}} \\ \cline{2-7}
                                  & \textbf{HR-Comp.} & \textbf{HR-Pres.} & \textbf{LR-Comp.} & \textbf{LR-Pres.} & \multicolumn{1}{c|}{\textbf{LR-R-U}} & \textbf{Avg.}  &                                                                                   \\ \hline
LLaVA-1.5~\cite{llava15}          & 67.30             & 69.80             & 68.20             & 55.50             & \multicolumn{1}{c|}{59.00}           & 63.96          & -                                                                                 \\
LLaVA-1.6~\cite{scorers}          & 68.60             & 64.40             & 64.32             & 56.84             & \multicolumn{1}{c|}{61.00}           & 63.03          & -                                                                                 \\
Qwen2-VL~\cite{qwen2vl}           & 75.60             & 63.30             & 75.47             & 62.00             & \multicolumn{1}{c|}{73.00}           & 69.87          & -                                                                                 \\
Qwen2.5-VL~\cite{qwen25vl}        & 75.28             & 67.30             & 73.86             & 64.67             & \multicolumn{1}{c|}{66.00}           & 69.42          & 51.21                                                                             \\
Qwen3-VL~\cite{qwen25vl}           & 81.00             & 78.10             & 70.32             & 56.42             & \multicolumn{1}{c|}{72.00}           & 71.57          & 54.75                                                                             \\
InternVL-2.5~\cite{internvl25}    & 75.50             & 65.80             & 71.16             & 66.21             & \multicolumn{1}{c|}{72.00}           & 70.13          & 47.20                                                                             \\
InternVL3~\cite{internvl3}        & 74.15             & 62.35             & 73.06             & 66.23             & \multicolumn{1}{c|}{74.00}           & 69.96          & 50.68                                                                             \\
InternVL3.5~\cite{internvl35}     & 80.14             & 53.80             & 92.00             & 91.26             & \multicolumn{1}{c|}{96.00}           & 82.64          & 53.74                                                                             \\
Mimo-VL~\cite{mimovl}      & 66.00             & 77.80             & 74.42             & 59.98             & \multicolumn{1}{c|}{65.00}           & 68.64          & 48.37                                                                             \\
LHRS-Bot-Nova~\cite{lhrsbot-nova} & 89.30             & 87.60             & 88.11             & 83.89             & \multicolumn{1}{c|}{79.00}           & 85.58          & -                                                                                 \\
GeoChat~\cite{geochat}            & 83.30             & 59.10             & 90.52             & 90.63             & \multicolumn{1}{c|}{97.00}           & 84.11          & 40.80                                                                             \\
VHM~\cite{vhm}                    & 83.30             & 68.30             & 90.11             & 89.89             & \multicolumn{1}{c|}{87.00}           & 83.72          & -                                                                                 \\ \hline
\textbf{RSCoVLM}                  & 82.60             & 68.50             & 93.16             & 92.18             & \multicolumn{1}{c|}{94.00}           & \textbf{86.09} & \textbf{58.08}                                                                    \\ \hline
\end{tabular}
\end{table*}

\subsubsection{Benchmark, Metric, and Comparison Setting}

We selected the most widely-used aerial image object detection benchmark, DOTA-v1.0~\cite{dota}, for our evaluation. The whole DOTA-v1.0 dataset comprises 2,806 high-resolution aerial images and 188,282 object instances across 15 common categories. The proportions of testing set is 1/3. These images were collected from multiple sensors and platforms, and each instance is annotated with a 8 degrees-of-freedom oriented bounding box, capturing the wide variations in object scale, shape, and orientation typical of aerial imagery.

We adopt the Average Precision with no confidence ($\text{AP}_{\text{nc}}$) and report three specific variants: $\text{AP}_{\text{nc}50}$ (IoU threshold is 0.50), $\text{AP}_{\text{nc}75}$ (IoU threshold is 0.75), and $\text{AP}_{\text{nc}50:95}$ (the average $\text{AP}_{\text{nc}}$ computed over IoU thresholds from 0.50 to 0.95 at increments of 0.05). The evaluation is based on the standard MMRotate~\cite{mmrotate} evaluation procedure. And the splitting length is set 512 with an overlap of 100.

The conventional object detection baselines are trained using the latest MMRotate~\cite{mmrotate}, and the details necessary for reproducibility are also provided in the released code. We obtain a reasonable $\text{AP}_{\text{nc}}$ for comparison methods using the following procedure: We first select a threshold for confidence scores to filter out low-score predictions, and then randomize (or, set to 1) the remaining prediction scores. The AP computed under this condition is denoted as $\text{AP}_{\text{nc}}$ of the conventional detector. To determine an appropriate threshold for each detector, we evaluate $\text{AP}_{\text{nc}}$ on the validation set by varying the confidence threshold from 0.00 to 0.95 in increments of 0.05, and select the threshold that yields the highest $\text{AP}_{\text{nc}}$ for subsequent evaluation on the test set. 

\subsubsection{Results}

We compare our model with state-of-the-art RS object detection methods. Our multi-task model achieves detection performance comparable to conventional detectors, even though it is not specifically optimized for the single dataset as the comparison methods are. When trained solely on object detection data, denoted as RSCoVLM-det, the model exhibits further improvement and even surpasses half of the conventional methods. It is a remarkable achievement for RS vision-language models.

Thanks to the dynamic resolution strategy, our model can further enhance detection performance by maximizing the inference scale, referred to as the “Max Mode.” Specifically, each input image is upsampled to the model’s upper input limit of 1008$\times$1008, and the outputs are then downsampled back to the original scale for evaluation. We observe a substantial increase in overall $\text{AP}_{\text{nc}}$, although certain categories such as plane (PL) and bridge (BD) experience minor degradation. The enhanced RSCoVLM-det even outperforms all competing approaches, while the conventional detectors are trained and evaluated at fixed resolutions without such a feature of test-time augmentation.

To the best of our knowledge, only two existing RS VLMs, LMMRotate~\cite{lmmrotate} (our conference version) and Falcon~\cite{falcon}, are capable of performing aerial object detection effectively. However, their common foundation model, Florence-2~\cite{florence2}, employs a fixed and relatively large input size of 1024$\times$1024, which already exceeds the input limit of RSCoVLM. Moreover, LMMRotate is trained specifically for detection, while Falcon performs well only on its training set and does not report test results. In addition, Falcon requires multiple inferences per image, making separate predictions for each category, which results in extremely high computational cost. Therefore, we consider RSCoVLM to be the only vision-language model capable of performing multiple tasks while achieving detection performance that is fairly comparable to specialized object detection models, currently.

\subsection{Evaluation on Scene Classification}

\subsubsection{Benchmark and Metric}

We evaluate our model on five standard remote‑sensing scene‑classification benchmarks. The AID~\cite{aid} dataset comprises approximately 10,000 images of size 600$\times$600 pixels across 30 classes. The UCMerced~\cite{ucm} dataset consists of 2,100 images of size 256$\times$256 pixels covering 21 classes. The NWPU‑RESISC45~\cite{resisc} dataset contains 31,500 images of size 256$\times$256 pixels across 45 classes, with large variation in resolution and scene complexity. The WHU‑RS19~\cite{whurs19} dataset includes around 1,000 high‑resolution patches of size 600$\times$600 pixels spanning 19 classes. The METER‑ML~\cite{meterml} benchmark offers a large‑scale multi‑sensor setup with varied image sizes for extended generalisation evaluation. Together these benchmarks allow a robust assessment of our model’s generalisation across dataset scale, class‑set size, imaging conditions and spatial resolutions.

We report overall accuracy of the test set for each benchmark. For METER‑ML, NWPU‑RESISC45, and WHU‑RS19, we adopt the test set splits defined by VHM~\cite{vhm}. For UCMerced, we follow the split defined by GeoChat~\cite{geochat}. For AID, we present results using both the VHM and GeoChat splits to facilitate fair comparison.

\subsubsection{Results}

Table~\ref{tab:res-cls} presents the comparative results across the five scene classification benchmarks. The compared methods include classical VLM baselines (MiniGPTv2~\cite{minigptv2} and LLaVA-1.5~\cite{llava15}), leading open-source VLMs (the QwenVL~\cite{qwen25vl}, InternVL~\cite{internvl25}, and MiMo-VL~\cite{mimovl} series), and latest RS VLMs (GeoChat~\cite{geochat}, TEOChat~\cite{irvin2024teochat}, LHRS-Bot-Nova~\cite{lhrsbot-nova}, SkysenseGPT~\cite{luo2024skysensegpt}, VHM~\cite{vhm}, and ScoreRS~\cite{scorers}). As shown, our model consistently surpasses all compared approaches across all benchmarks.

\subsection{Evaluation on Visual Question Answering}

\subsubsection{Benchmark and Metric}

We evaluate our model’s visual question answering capability using two established benchmarks in the RS domain, including the RSVQA benchmark~\cite{rsvqa} and the VQA portion of VRSBench~\cite{vrsbench}. The RSVQA compromise two subsets of image-question-answer triplet derived from high‑resolution (HR) orthorectified imagery and low‑resolution (LR) RS data, enabling evaluation of model reasoning across spatial scales. The VRSBench dataset is a large‑scale vision‑language benchmark for RS image understanding that comprises 37,408 question‑answer pairs in test set, supporting a broad range of understanding instructions. The standard question answering accuracy is used as the metric.

\subsubsection{Results}

Table~\ref{tab:res-vqa} presents the results of visual question answering, demonstrating the strong understanding and conversational capabilities of our model. Our approach surpasses all open-source VLMs (including the LLaVA, Qwen, InternVL, and MiMo-VL series) as well as RS VLMs (GeoChat~\cite{geochat}, LHRS-Bot-Nova~\cite{lhrsbot-nova}, and VHM~\cite{vhm}) across the two benchmarks. In the zero-shot question answering evaluation on VRSBench-VQA~\cite{vrsbench}, our model outperforms the latest general-purpose models, showing superior generalization ability on RS image question answering.

\section{Conclusion}

In this paper, we introduce RSCoVLM, the latest generation of versatile vision language model. We carefully curated RS data, detailing the processes of data collection, offline integration, and online loading with adaptive weighting. To handle the wide range of image resolutions in RS images, we developed a dynamic-resolution strategy and proposed the Zoom-in Chain mechanism with the LRS-VQA-Zoom dataset for ultra-high-resolution images. Moreover, we improved the model’s object detection capabilities and designed a fair evaluation protocol for comparison with conventional methods. 
Comprehensive experiments show that RSCoVLM consistently delivers state-of-the-art results across multiple tasks, surpassing previous RS VLMs and matching task-specific expert models. By releasing all code, models, and datasets, we aim to enable reproducibility and foster progress toward general-purpose remote sensing models.

\bibliographystyle{IEEEtran}
\bibliography{reference}

@article{TRD,
	title={Transformer with Transfer CNN for Remote-Sensing-Image Object Detection},
	author={Li, Qingyun and Chen, Yushi and Zeng, Ying},
	journal={Remote Sens.},
	volume={14},
	number={4},
	pages={984},
	year={2022},
	publisher={MDPI}
}

@inproceedings{ViT,
	title={An Image is Worth 16x16 Words: Transformers for Image Recognition at Scale},
	author={Alexey Dosovitskiy and Lucas Beyer and Alexander Kolesnikov and others},
	booktitle={ICLR},
	year={2021}
}

@article{Transformer,
	title={Attention is all you need},
	author={Vaswani, Ashish and Shazeer, Noam and Parmar, Niki and others},
	journal={NIPS},
	volume={30},
	year={2017}
}

@ARTICLE{resisc,  
	author={Cheng, Gong and Han, Junwei and Lu, Xiaoqiang},
	journal={Proc. of the IEEE},
	title={Remote Sensing Image Scene Classification: Benchmark and State of the Art},
	year={2017},
	volume={105},
	number={10},
	pages={1865-1883},
	doi={10.1109/JPROC.2017.2675998}
}

@ARTICLE{aid,
	author={Xia, Gui-Song and Hu, Jingwen and Hu, Fan and others},
	journal={IEEE Trans. Geosci. Remote Sens.}, 
	title={{AID}: A Benchmark Data Set for Performance Evaluation of Aerial Scene Classification}, 
	year={2017},
	volume={55},
	number={7},
	pages={3965-3981},
	doi={10.1109/TGRS.2017.2685945}
}

@InProceedings{dota,
	author = {Xia, Gui-Song and Bai, Xiang and Ding, Jian and others},
	title = {{DOTA}: A Large-Scale Dataset for Object Detection in Aerial Images},
	booktitle = {CVPR},
	month = {June},
	year = {2018}
}

@ARTICLE{mtl_survey_2022,
	author={Zhang, Yu and Yang, Qiang},
	journal={IEEE Trans. on Know. and Data Engin.}, 
	title={A Survey on Multi-Task Learning}, 
	year={2022},
	volume={34},
	number={12},
	pages={5586-5609},
	doi={10.1109/TKDE.2021.3070203}
}

@article{FasterRCNN,
	title={Faster r-cnn: Towards real-time object detection with region proposal networks},
	author={Ren, Shaoqing and He, Kaiming and Girshick, Ross and Sun, Jian},
	journal={NIPS},
	volume={28},
	year={2015}
}

@ARTICLE{GRSM22review,
	author={Zhang, Lefei and Zhang, Liangpei},
	journal={IEEE Geosci. Remote Sens. Magaz.}, 
	title={Artificial Intelligence for Remote Sensing Data Analysis: A review of challenges and opportunities}, 
	year={2022},
	volume={10},
	number={2},
	pages={270-294},
	doi={10.1109/MGRS.2022.3145854}
}

@inproceedings{clip,
  title={Learning Transferable Visual Models From Natural Language Supervision},
  author={Alec Radford and Jong Wook Kim and Chris Hallacy and A. Ramesh and Gabriel Goh and Sandhini Agarwal and Girish Sastry and Amanda Askell and Pamela Mishkin and Jack Clark and Gretchen Krueger and Ilya Sutskever},
  booktitle={ICML},
  year={2021}
}

@inproceedings{r3det,
  title={R3det: Refined single-stage detector with feature refinement for rotating object},
  author={Yang, Xue and Yan, Junchi and Feng, Ziming and He, Tao},
  booktitle={AAAI},
  year={2021}
}

@inproceedings{csl,
  title={Arbitrary-oriented object detection with circular smooth label},
  author={Yang, Xue and Yan, Junchi},
  booktitle={ECCV},
  year={2020},
}

@InProceedings{visualgpt,
    author    = {Chen, Jun and Guo, Han and Yi, Kai and others},
    title     = {VisualGPT: Data-Efficient Adaptation of Pretrained Language Models for Image Captioning},
    booktitle = {Proceedings of the IEEE/CVF Conference on Computer Vision and Pattern Recognition (CVPR)},
    month     = {June},
    year      = {2022},
    pages     = {18030-18040}
}

@InProceedings{blip2,
  title = 	 {{BLIP}-2: Bootstrapping Language-Image Pre-training with Frozen Image Encoders and Large Language Models},
  author =       {Li, Junnan and Li, Dongxu and Savarese, Silvio and Hoi, Steven},
  booktitle = 	 {Proceedings of the 40th International Conference on Machine Learning},
  pages = 	 {19730--19742},
  year = 	 {2023},
  editor = 	 {Krause, Andreas and Brunskill, Emma and Cho, Kyunghyun and Engelhardt, Barbara and Sabato, Sivan and Scarlett, Jonathan},
  volume = 	 {202},
  series = 	 {Proceedings of Machine Learning Research},
  month = 	 {23--29 Jul},
  publisher =    {PMLR},
}

@inproceedings{flamingo,
 author = {Alayrac, Jean-Baptiste and Donahue, Jeff and Luc, Pauline and others},
 booktitle = {Advances in Neural Information Processing Systems},
 pages = {23716--23736},
 publisher = {Curran Associates, Inc.},
 title = {Flamingo: a Visual Language Model for Few-Shot Learning},
 volume = {35},
 year = {2022}
}

@inproceedings{llava,
 author = {Liu, Haotian and Li, Chunyuan and Wu, Qingyang and Lee, Yong Jae},
 booktitle = {Advances in Neural Information Processing Systems},
 editor = {A. Oh and T. Naumann and A. Globerson and K. Saenko and M. Hardt and S. Levine},
 pages = {34892--34916},
 publisher = {Curran Associates, Inc.},
 title = {Visual Instruction Tuning},
 volume = {36},
 year = {2023}
}

@inproceedings{zhu2024minigpt,
title={Mini{GPT}-4: Enhancing Vision-Language Understanding with Advanced Large Language Models},
author={Deyao Zhu and Jun Chen and Xiaoqian Shen and Xiang Li and Mohamed Elhoseiny},
booktitle={The Twelfth International Conference on Learning Representations},
year={2024},
}

@inproceedings{InstructBlip,
 author = {Dai, Wenliang and Li, Junnan and LI, DONGXU and others},
 booktitle = {Advances in Neural Information Processing Systems},
 editor = {A. Oh and T. Naumann and A. Globerson and K. Saenko and M. Hardt and S. Levine},
 pages = {49250--49267},
 publisher = {Curran Associates, Inc.},
 title = {InstructBLIP: Towards General-purpose Vision-Language Models with Instruction Tuning},
 volume = {36},
 year = {2023}
}

@misc{lamaadapterv2,
      title={LLaMA-Adapter V2: Parameter-Efficient Visual Instruction Model}, 
      author={Peng Gao and Jiaming Han and Renrui Zhang and others},
      year={2023},
      eprint={2304.15010},
      archivePrefix={arXiv},
      primaryClass={cs.CV},
}

@misc{sphinx,
      title={SPHINX: The Joint Mixing of Weights, Tasks, and Visual Embeddings for Multi-modal Large Language Models}, 
      author={Ziyi Lin and Chris Liu and Renrui Zhang and others},
      year={2023},
      eprint={2311.07575},
      archivePrefix={arXiv},
      primaryClass={cs.CV},
}

@inproceedings{VisionLLM,
 author = {Wang, Wenhai and Chen, Zhe and Chen, Xiaokang and others},
 booktitle = {Advances in Neural Information Processing Systems},
 editor = {A. Oh and T. Naumann and A. Globerson and K. Saenko and M. Hardt and S. Levine},
 pages = {61501--61513},
 publisher = {Curran Associates, Inc.},
 title = {VisionLLM: Large Language Model is also an Open-Ended Decoder for Vision-Centric Tasks},
 volume = {36},
 year = {2023}
}

@inproceedings{visionllm-v2,
 author = {Wu, Jiannan and Zhong, Muyan and Xing, Sen and others},
 booktitle = {Advances in Neural Information Processing Systems},
 editor = {A. Globerson and L. Mackey and D. Belgrave and A. Fan and U. Paquet and J. Tomczak and C. Zhang},
 pages = {69925--69975},
 publisher = {Curran Associates, Inc.},
 title = {VisionLLM v2: An End-to-End Generalist Multimodal Large Language Model for Hundreds of Vision-Language Tasks},
 volume = {37},
 year = {2024}
}

@misc{palix,
      title={PaLI-X: On Scaling up a Multilingual Vision and Language Model}, 
      author={Xi Chen and Josip Djolonga and Piotr Padlewski and others},
      year={2023},
      eprint={2305.18565},
      archivePrefix={arXiv},
      primaryClass={cs.CV},
}

@misc{mimovl,
      title={MiMo-VL Technical Report}, 
      author={Zihao Yue and Zhenru Lin and Yifan Song and others},
      year={2025},
      eprint={2506.03569},
      archivePrefix={arXiv},
      primaryClass={cs.CL},
}

@inproceedings{cogvlm,
 author = {Wang, Weihan and Lv, Qingsong and Yu, Wenmeng and others},
 booktitle = {Advances in Neural Information Processing Systems},
 editor = {A. Globerson and L. Mackey and D. Belgrave and A. Fan and U. Paquet and J. Tomczak and C. Zhang},
 pages = {121475--121499},
 publisher = {Curran Associates, Inc.},
 title = {CogVLM: Visual Expert for Pretrained Language Models},
 volume = {37},
 year = {2024}
}

@misc{qwen2vl,
      title={Qwen2-VL: Enhancing Vision-Language Model's Perception of the World at Any Resolution}, 
      author={Peng Wang and Shuai Bai and Sinan Tan and others},
      year={2024},
      eprint={2409.12191},
      archivePrefix={arXiv},
      primaryClass={cs.CV},
}

@misc{qwen25vl,
      title={Qwen2.5-VL Technical Report}, 
      author={Shuai Bai and Keqin Chen and Xuejing Liu and others},
      year={2025},
      eprint={2502.13923},
      archivePrefix={arXiv},
      primaryClass={cs.CV},
}

@InProceedings{geochat,
    author    = {Kuckreja, Kartik and Danish, Muhammad Sohail and Naseer, Muzammal and others},
    title     = {GeoChat: Grounded Large Vision-Language Model for Remote Sensing},
    booktitle = {Proceedings of the IEEE/CVF Conference on Computer Vision and Pattern Recognition (CVPR)},
    month     = {June},
    year      = {2024},
    pages     = {27831-27840}
}

@ARTICLE{earthgpt,
  author={Zhang, Wei and Cai, Miaoxin and Zhang, Tong and others},
  journal={IEEE Transactions on Geoscience and Remote Sensing}, 
  title={EarthGPT: A Universal Multimodal Large Language Model for Multisensor Image Comprehension in Remote Sensing Domain}, 
  year={2024},
  volume={62},
  number={},
  pages={1-20},
  doi={10.1109/TGRS.2024.3409624}}

@inproceedings{lhrsbot,
  title={Lhrs-bot: Empowering remote sensing with vgi-enhanced large multimodal language model},
  author={Muhtar, Dilxat and Li, Zhenshi and Gu, Feng and others},
  booktitle={European Conference on Computer Vision},
  pages={440--457},
  year={2024},
  organization={Springer}
}

@article{irvin2024teochat,
  title={Teochat: A large vision-language assistant for temporal earth observation data},
  author={Irvin, Jeremy Andrew and Liu, Emily Ruoyu and Chen, Joyce Chuyi and others},
  journal={arXiv preprint arXiv:2410.06234},
  year={2024}
}

@article{luo2024skysensegpt,
  title={Skysensegpt: A fine-grained instruction tuning dataset and model for remote sensing vision-language understanding},
  author={Luo, Junwei and Pang, Zhen and Zhang, Yongjun and others},
  journal={arXiv preprint arXiv:2406.10100},
  year={2024}
}

@inproceedings{vhm,
  title={Vhm: Versatile and honest vision language model for remote sensing image analysis},
  author={Pang, Chao and Weng, Xingxing and Wu, Jiang and others},
  booktitle={Proceedings of the AAAI Conference on Artificial Intelligence},
  volume={39},
  number={6},
  pages={6381--6388},
  year={2025}
}

@article{MTSCD-Net,
title = {MTSCD-Net: A network based on multi-task learning for semantic change detection of bitemporal remote sensing images},
journal = {International Journal of Applied Earth Observation and Geoinformation},
volume = {118},
pages = {103294},
year = {2023},
issn = {1569-8432},
author = {Fengzhi Cui and Jie Jiang},
}

@article{stgnet,
  title={Multitask semantic change detection guided by spatiotemporal semantic interaction},
  author={Wang, Yinqing and Zhao, Liangjun and Hu, Yueming and Dai, Hui and Zhang, Yuanyang},
  journal={Scientific Reports},
  volume={15},
  number={1},
  pages={16003},
  year={2025},
  publisher={Nature Publishing Group UK London}
}

@article{SMNet,
  title={SMNet: Symmetric multi-task network for semantic change detection in remote sensing images based on CNN and transformer},
  author={Niu, Yiting and Guo, Haitao and Lu, Jun and Ding, Lei and Yu, Donghang},
  journal={Remote Sensing},
  volume={15},
  number={4},
  pages={949},
  year={2023},
  publisher={MDPI}
}

@article{mcenet,
  title={A multi-task consistency enhancement network for semantic change detection in HR remote sensing images and application of non-agriculturalization},
  author={Lin, Haihan and Wang, Xiaoqin and Li, Mengmeng and Huang, Dehua and Wu, Ruijiao},
  journal={Remote Sensing},
  volume={15},
  number={21},
  pages={5106},
  year={2023},
  publisher={MDPI}
}

@article{lu2022multi,
  title={Multi-task learning of relative height estimation and semantic segmentation from single airborne RGB images},
  author={Lu, Min and Liu, Jiayin and Wang, Feng and Xiang, Yuming},
  journal={Remote Sensing},
  volume={14},
  number={14},
  pages={3450},
  year={2022},
  publisher={MDPI}
}

@article{carvalho2019multitask,
  title={Multitask learning of height and semantics from aerial images},
  author={Carvalho, Marcela and Le Saux, Bertrand and Trouv{\'e}-Peloux, Pauline and Champagnat, Fr{\'e}d{\'e}ric and Almansa, Andr{\'e}s},
  journal={IEEE Geoscience and Remote Sensing Letters},
  volume={17},
  number={8},
  pages={1391--1395},
  year={2019},
  publisher={IEEE}
}

@article{li2024co,
  title={Co-training transformer for remote sensing image classification, segmentation, and detection},
  author={Li, Qingyun and Chen, Yushi and He, Xin and Huang, Lingbo},
  journal={IEEE Transactions on Geoscience and Remote Sensing},
  volume={62},
  pages={1--18},
  year={2024},
  publisher={IEEE}
}

@InProceedings{Bastani_2023_ICCV,
    author    = {Bastani, Favyen and Wolters, Piper and Gupta, Ritwik and Ferdinando, Joe and Kembhavi, Aniruddha},
    title     = {SatlasPretrain: A Large-Scale Dataset for Remote Sensing Image Understanding},
    booktitle = {Proceedings of the IEEE/CVF International Conference on Computer Vision (ICCV)},
    month     = {October},
    year      = {2023},
    pages     = {16772-16782}
}

@article{li2024sm3det,
  title={Sm3det: A unified model for multi-modal remote sensing object detection},
  author={Li, Yuxuan and Li, Xiang and Li, Yunheng and others},
  journal={arXiv preprint arXiv:2412.20665},
  year={2024}
}

@inproceedings{soni2025earthdial,
  title={Earthdial: Turning multi-sensory earth observations to interactive dialogues},
  author={Soni, Sagar and Dudhane, Akshay and Debary, Hiyam and others},
  booktitle={Proceedings of the Computer Vision and Pattern Recognition Conference},
  pages={14303--14313},
  year={2025}
}

@article{zhou2024vlgfm,
  title={Towards Vision-Language Geo-Foundation Models: A Survey},
  author={Yue Zhou and Litong Feng and Yiping Ke and others},
  journal={arXiv preprint arXiv:2406.09385},
  year={2024}
}

@article{han2023onellm,
  title={OneLLM: One Framework to Align All Modalities with Language}, 
  author={Han, Jiaming and Gong, Kaixiong and Zhang, Yiyuan and others},
  year={2023},
  eprint={2312.03700},
  archivePrefix={arXiv},
  primaryClass={cs.CV}
}

@inproceedings{li2024omnicorpus,
  title={OmniCorpus: A Unified Multimodal Corpus of 10 Billion-Level Images Interleaved with Text},
  author={Li, Qingyun and Chen, Zhe and Wang, Weiyun and others},
  booktitle={The Thirteenth International Conference on Learning Representations},
  year={2025}
}

@ARTICLE{MTP,
  author={Wang, Di and Zhang, Jing and Xu, Minqiang and others},
  journal={IEEE Journal of Selected Topics in Applied Earth Observations and Remote Sensing}, 
  title={MTP: Advancing Remote Sensing Foundation Model Via Multi-Task Pretraining}, 
  year={2024},
  volume={},
  number={},
  pages={1-24},
  doi={10.1109/JSTARS.2024.3408154}
}

@article{lrsvqa,
    title={When Large Vision-Language Model Meets Large Remote Sensing Imagery: Coarse-to-Fine Text-Guided Token Pruning},
    author={Luo, Junwei and Zhang, Yingying and Yang, Xue and others},
    journal={arXiv preprint arXiv:2503.07588},
    year={2025}
}

@article{geollava8k,
      title={GeoLLaVA-8K: Scaling Remote-Sensing Multimodal Large Language Models to 8K Resolution}, 
      author={Fengxiang Wang and Mingshuo Chen and Yueying Li and others},
  journal={arXiv preprint arXiv:2505.21375},
      year={2025},
}

@article{lmmrotate,
  title={A Simple Aerial Detection Baseline of Multimodal Language Models},
  author={Li, Qingyun and Chen, Yushi and Shu, Xinya and others},
  journal={arXiv preprint arXiv:2501.09720},
  year={2025}
}

@misc{geoground,
      title={GeoGround: A Unified Large Vision-Language Model for Remote Sensing Visual Grounding}, 
      author={Yue Zhou and Mengcheng Lan and Xiang Li and others},
      year={2024},
      eprint={2411.11904},
      archivePrefix={arXiv},
      primaryClass={cs.CV},
}

@article{
    li2025llavaonevision,
    title={{LL}a{VA}-OneVision: Easy Visual Task Transfer},
    author={Bo Li and Yuanhan Zhang and Dong Guo and others},
    journal={Transactions on Machine Learning Research},
    issn={2835-8856},
    year={2025},
    note={}
}

@article{li2024learning,
    title={Learning to Holistically Detect Bridges From Large-Size VHR Remote Sensing Imagery},
    author={Li, Yansheng and Luo, Junwei and Zhang, Yongjun and others},
    journal={IEEE Transactions on Pattern Analysis and Machine Intelligence},
    volume={44},
    number={11},
    pages={7778--7796},
    year={2024},
    publisher={IEEE}
}

@article{li2024scene,
   title={STAR: A First-Ever Dataset and A Large-Scale Benchmark for Scene Graph Generation in Large-Size Satellite Imagery},
   author={Li, Yansheng and Wang, Linlin and Wang, Tingzhu and others},
   journal={arXiv preprint arXiv:2406.09410},
   year={2024}
}

@article{FAIR1M,
    title = {FAIR1M: A benchmark dataset for fine-grained object recognition in high-resolution remote sensing imagery},
    journal = {ISPRS J. Photogram. Remote Sens.},
    volume = {184},
    pages = {116-130},
    year = {2022},
    issn = {0924-2716},
    author = {Xian Sun and Peijin Wang and Zhiyuan Yan and others},
}

@inproceedings{zero,
author = {Rajbhandari, Samyam and Rasley, Jeff and Ruwase, Olatunji and He, Yuxiong},
title = {ZeRO: memory optimizations toward training trillion parameter models},
year = {2020},
isbn = {9781728199986},
publisher = {IEEE Press},
booktitle = {Proceedings of the International Conference for High Performance Computing, Networking, Storage and Analysis},
articleno = {20},
numpages = {16},
location = {Atlanta, Georgia},
series = {SC '20}
}

@inproceedings{
dao2024flashattention,
title={FlashAttention-2: Faster Attention with Better Parallelism and Work Partitioning},
author={Tri Dao},
booktitle={The Twelfth International Conference on Learning Representations},
year={2024},
}

@inproceedings{
hsu2025ligerkernel,
title={Liger-Kernel: Efficient Triton Kernels for {LLM} Training},
author={Pin-Lun Hsu and Yun Dai and Vignesh Kothapalli and others},
booktitle={Championing Open-source DEvelopment in ML Workshop @ ICML25},
year={2025},
}

@inproceedings{vllm,
author = {Kwon, Woosuk and Li, Zhuohan and Zhuang, Siyuan and others},
title = {Efficient Memory Management for Large Language Model Serving with PagedAttention},
year = {2023},
isbn = {9798400702297},
publisher = {Association for Computing Machinery},
address = {New York, NY, USA},
doi = {10.1145/3600006.3613165},
booktitle = {Proceedings of the 29th Symposium on Operating Systems Principles},
pages = {611–626},
numpages = {16},
location = {Koblenz, Germany},
series = {SOSP '23}
}

@ARTICLE{rsvg,
  author={Zhan, Yang and Xiong, Zhitong and Yuan, Yuan},
  journal={IEEE Transactions on Geoscience and Remote Sensing}, 
  title={RSVG: Exploring Data and Models for Visual Grounding on Remote Sensing Data}, 
  year={2023},
  volume={61},
  number={},
  pages={1-13},
  doi={10.1109/TGRS.2023.3250471}
}

@article{vrsbench,
  title={VRSBench: A Versatile Vision-Language Benchmark Dataset for Remote Sensing Image Understanding},
  author={Xiang Li and Jian Ding and Mohamed Elhoseiny},
  journal={arXiv:2406.12384},
  year={2024}
}

@inproceedings{mmrotate,
  title   = {MMRotate: A Rotated Object Detection Benchmark using PyTorch},
  author  = {Zhou, Yue and Yang, Xue and Zhang, Gefan and others},
  booktitle={Proceedings of the 30th ACM International Conference on Multimedia},
  year={2022}
}

@article{falcon,
  title={Falcon: A Remote Sensing Vision-Language Foundation Model},
  author={kelu, Yao and Nuo, Xu and Rong, Yang and others},
  journal={arXiv preprint arXiv:2503.11070},
  year={2025}
}

@article{florence2,
  title={Florence-2: Advancing a unified representation for a variety of vision tasks},
  author={Xiao, Bin and Wu, Haiping and Xu, Weijian and others},
  journal={arXiv preprint arXiv:2311.06242},
  year={2023}
}

@inproceedings{ucm,
author = {Yang, Yi and Newsam, Shawn},
title = {Bag-of-visual-words and spatial extensions for land-use classification},
year = {2010},
isbn = {9781450304283},
publisher = {Association for Computing Machinery},
address = {New York, NY, USA},
doi = {10.1145/1869790.1869829},
booktitle = {Proceedings of the 18th SIGSPATIAL International Conference on Advances in Geographic Information Systems},
pages = {270–279},
numpages = {10},
location = {San Jose, California},
series = {GIS '10}
}

@Article{whurs19,
title={Satellite Image Classification via Two-Layer Sparse Coding With Biased Image Representation},
author={Dengxin Dai and Wen Yang},
journal={IEEE Transactions on Geoscience and Remote Sensing},
year={2011},
volume={8},
number={1},
pages={173-176}
}

@article{meterml,
  title={METER-ML: a multi-sensor earth observation benchmark for automated methane source mapping},
  author={Zhu, Bryan and Lui, Nicholas and Irvin, Jeremy and others},
  journal={arXiv preprint arXiv:2207.11166},
  year={2022}
}

@article{rsvqa,
  title={RSVQA: Visual question answering for remote sensing data},
  author={Lobry, Sylvain and Marcos, Diego and Murray, Jesse and Tuia, Devis},
  journal={IEEE Transactions on Geoscience and Remote Sensing},
  volume={58},
  number={12},
  pages={8555--8566},
  year={2020},
  publisher={IEEE}
}

@article{minigptv2,
  title={MiniGPT-v2: large language model as a unified interface for vision-language multi-task learning}, 
  author={Chen, Jun and Zhu, Deyao and Shen, Xiaoqian and others},
  year={2023},
  journal={arXiv preprint arXiv:2310.09478},
}

@misc{llava15,
      title={Improved Baselines with Visual Instruction Tuning}, 
      author={Liu, Haotian and Li, Chunyuan and Li, Yuheng and Lee, Yong Jae},
      publisher={arXiv:2310.03744},
      year={2023},
}

@article{internvl35,
  title={InternVL3.5: Advancing Open-Source Multimodal Models in Versatility, Reasoning, and Efficiency},
  author={Wang, Weiyun and Gao, Zhangwei and Gu, Lixin and others},
  journal={arXiv preprint arXiv:2508.18265},
  year={2025}
}

@article{internvl3,
  title={Internvl3: Exploring advanced training and test-time recipes for open-source multimodal models},
  author={Zhu, Jinguo and Wang, Weiyun and Chen, Zhe and others},
  journal={arXiv preprint arXiv:2504.10479},
  year={2025}
}

@article{internvl25,
  title={Expanding Performance Boundaries of Open-Source Multimodal Models with Model, Data, and Test-Time Scaling},
  author={Chen, Zhe and Wang, Weiyun and Cao, Yue and others},
  journal={arXiv preprint arXiv:2412.05271},
  year={2024}
}

@article{mpo,
  title={Enhancing the Reasoning Ability of Multimodal Large Language Models via Mixed Preference Optimization},
  author={Wang, Weiyun and Chen, Zhe and Wang, Wenhai and others},
  journal={arXiv preprint arXiv:2411.10442},
  year={2024}
}

@inproceedings{internvl,
  title={Internvl: Scaling up vision foundation models and aligning for generic visual-linguistic tasks},
  author={Chen, Zhe and Wu, Jiannan and Wang, Wenhai and others},
  booktitle={Proceedings of the IEEE/CVF Conference on Computer Vision and Pattern Recognition},
  pages={24185--24198},
  year={2024}
}

@inproceedings{scorers,
title={Quality-Driven Curation of Remote Sensing Vision-Language Data via Learned Scoring Models},
author={Dilxat Muhtar and Enzhuo Zhang and Zhenshi Li and others},
booktitle={The Thirty-ninth Annual Conference on Neural Information Processing Systems},
year={2025},
}

@article{lhrsbot-nova,
  title={Lhrs-bot-nova: Improved multimodal large language model for remote sensing vision-language interpretation},
  author={Li, Zhenshi and Muhtar, Dilxat and Gu, Feng and others},
  journal={ISPRS Journal of Photogrammetry and Remote Sensing},
  volume={227},
  pages={539--550},
  year={2025},
  publisher={Elsevier}
}

@inproceedings{gwd,
  title={Rethinking rotated object detection with gaussian wasserstein distance loss},
  author={Yang, Xue and Yan, Junchi and Ming, Qi and others},
  booktitle={International conference on machine learning},
  pages={11830--11841},
  year={2021},
  organization={PMLR}
}

@inproceedings{atss,
  title={Bridging the gap between anchor-based and anchor-free detection via adaptive training sample selection},
  author={Zhang, Shifeng and Chi, Cheng and Yao, Yongqiang and others},
  booktitle={Proceedings of the IEEE/CVF conference on computer vision and pattern recognition},
  pages={9759--9768},
  year={2020}
}

@inproceedings{fcos,
  title={Fcos: Fully convolutional one-stage object detection},
  author={Tian, Zhi and Shen, Chunhua and Chen, Hao and He, Tong},
  booktitle={Proceedings of the IEEE/CVF international conference on computer vision},
  pages={9627--9636},
  year={2019}
}

@article{s2anet,
  title={Align deep features for oriented object detection},
  author={Han, Jiaming and Ding, Jian and Li, Jie and Xia, Gui-Song},
  journal={IEEE transactions on geoscience and remote sensing},
  volume={60},
  pages={1--11},
  year={2021},
  publisher={IEEE}
}

@article{llavauhdv2,
  title={LLaVA-UHD v2: an MLLM Integrating High-Resolution Semantic Pyramid via Hierarchical Window Transformer},
  author={Zhang, Yipeng and Liu, Yifan and Guo, Zonghao and Zhang, Yidan and Yang, Xuesong and Zhang, Xiaoying and Chen, Chi and Song, Jun and Zheng, Bo and Yao, Yuan and others},
  journal={arXiv preprint arXiv:2412.13871},
  year={2024}
}

\vfill

\end{document}